\pdfoutput=1
\documentclass[11pt]{article}

\usepackage[]{EMNLP2023}
\usepackage{times}
\usepackage{latexsym}

\usepackage[T1]{fontenc}
\usepackage[utf8]{inputenc}

\usepackage{microtype}

\usepackage{inconsolata}

\usepackage[utf8]{inputenc}
\usepackage{booktabs}
\usepackage{graphicx}
\usepackage{CJK}
\usepackage{multirow}
\usepackage{verbatim}
\usepackage{url}
\usepackage{enumitem}
\usepackage{amsmath}
\usepackage{diagbox}
\usepackage{flushend,cuted}
\usepackage{bbm}
\usepackage{xcolor}
\usepackage{xspace}
\usepackage{algorithm}
\usepackage{algpseudocode}
\usepackage{verbatim}
\usepackage{colortbl}
\usepackage{bbding}
\usepackage{tablefootnote}

\usepackage{listings}
\usepackage{ulem}
\usepackage{makecell}
\usepackage{float}
\usepackage{color}
\usepackage{makecell}

\title{Multilingual Machine Translation with Large Language Models: \\ Empirical Results and Analysis}
\author{
    Wenhao Zhu$^{1,2}$\text{, } \textbf{Hongyi Liu}$^{3}$\text{, } \textbf{Qingxiu Dong}$^{4}$\text{, } \textbf{Jingjing Xu}$^{2}$ \\
     \textbf{Shujian Huang}$^{1}$ \textbf{, } \textbf{Lingpeng Kong}$^{5}$\textbf{, } \textbf{Jiajun Chen}$^{1}$\textbf{, } \textbf{Lei Li}$^{6}$ \\
    $^{1}$ \text{National Key Laboratory for Novel Software Technology, Nanjing University} \\
    $^{2}$ \text{Shanghai AI Lab}  $^{3}$ \text{Shanghai Jiao Tong University} $^{4}$ \text{Peking University} \\
    $^{5}$ \text{The University of Hong Kong} $^{6}$ \text{Language Technologies Institute, Carnegie Mellon University} \\
    \small\texttt{zhuwh@smail.nju.edu.cn}, \small\texttt{liu.hong.yi@sjtu.edu.cn}, \small\texttt{dqx@stu.pku.edu.cn}, \small\texttt{jingjingxu@pku.edu.cn} \\
    \small\texttt{huangsj@nju.edu.cn}, \small\texttt{lpk@cs.hku.hk}, \small\texttt{chenjj@nju.edu.cn}, \small\texttt{leili@cs.cmu.edu} \\
}

\begin{document}
\maketitle

% \renewcommand{\thefootnote}{\fnsymbol{footnote}}
% \footnotetext[1]{Equal Contributions}
% \renewcommand{\thefootnote}{\arabic{footnote}}

\begin{abstract}
Large language models (LLMs) have demonstrated remarkable potential in handling multilingual machine translation (MMT). In this paper, we systematically investigate the advantages and challenges of LLMs for MMT by answering two questions: 1) How well do LLMs perform in translating massive languages? 2) Which factors affect LLMs' performance in translation? We thoroughly evaluate eight popular LLMs, including ChatGPT and GPT-4.
Our empirical results show that translation capabilities of LLMs are continually involving. 
GPT-4 has beat the strong supervised baseline NLLB in 40.91\% of translation directions but still faces a large gap towards the commercial translation system like Google Translate, especially on low-resource languages.
Through further analysis, we discover that LLMs exhibit new working patterns when used for MMT. 
First, LLM can acquire translation ability in a resource-efficient way and generate moderate translation even on zero-resource languages.
Second, instruction semantics can surprisingly be ignored when given in-context exemplars. 
Third, cross-lingual exemplars can provide better task guidance for low-resource translation than exemplars in the same language pairs\footnote{Code will be released at: \url{https://github.com/NJUNLP/MMT-LLM}.}.
\end{abstract}

\section{Introduction}
With the increasing scale of parameters and training corpus, large language models (LLMs) have gained a universal ability to handle a variety of tasks via in-context learning (ICL, \citealt{brown2020language}), which allows language models to perform tasks with a few given exemplars and human-written instructions as context.
One particular area where LLMs have shown outstanding potential is machine translation (MT).
Previous studies have shown the surprising performance of LLMs on high-resource bilingual translation, such as English-German translation~\cite{vilar2022prompting, zhang2022opt}, even if these models are not particularly optimized on multilingual data.

\begin{figure}[t]
    \centering
    \includegraphics[width=0.42\textwidth]{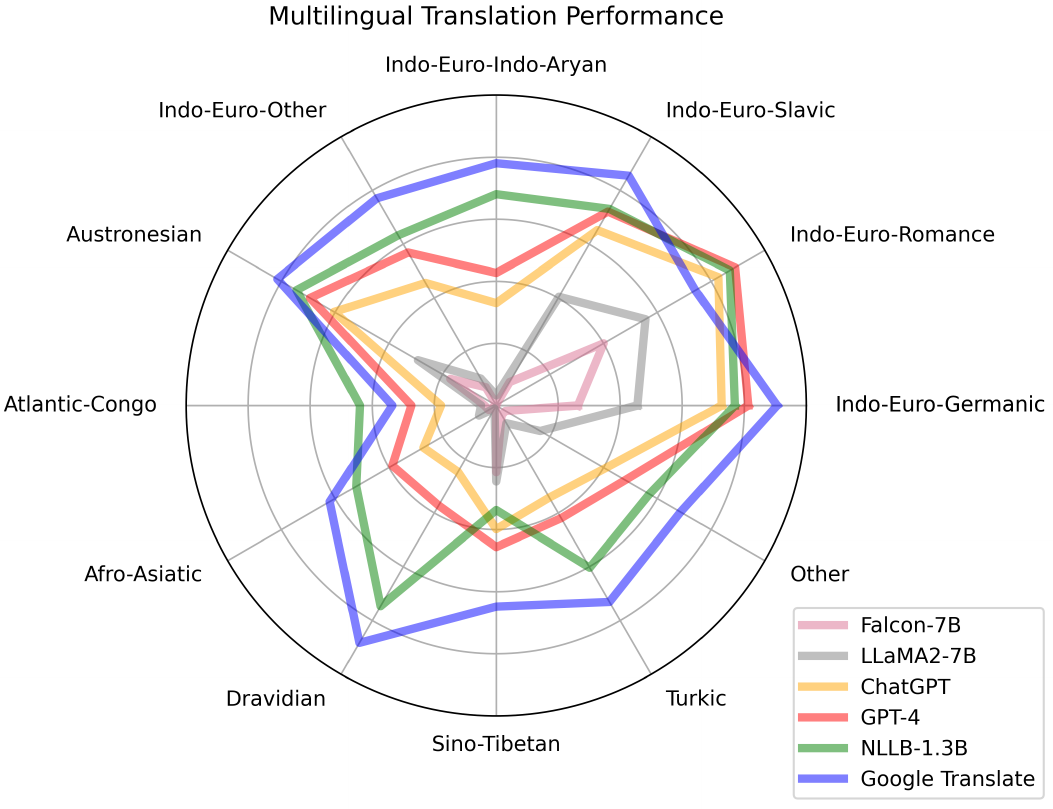} 
    \caption{Multilingual translation performance (BLEU) of some popular LLMs and traditional supervised systems in translating from English to non-English. LLMs have demonstrated great potential in multilingual machine translation.}
    \label{fig:llm}
\end{figure}

However, the multilingual translation ability of LLMs remains under-explored. 
MMT is a challenging task that involves translating text among different languages and requires semantic alignment between languages~\cite{fan2021beyond, costa2022no, yuan2023lego}.
It is also unclear that how LLM acquires translation ability and which factors affect LLM's translation ability.

In this paper, we follow ICL paradigm and focus on studying LLMs in multilingual machine translation by answering two questions: 
1) \textit{How LLMs perform MMT over massive languages?}  2) \textit{Which factors affect the performance of LLMs?}

For the first question, we evaluate several popular LLMs: English-centric LLMs, including OPT~\cite{zhang2022opt}, LLaMA2~\cite{touvron2023llama}, Falcon~\cite{falcon40b} and multilingual LLMs, including XGLM~\cite{lin2022few}, BLOOMZ~\cite{scao2022bloom}, ChatGPT~\cite{openai2022chatgpt}, GPT-4~\cite{openai2023gpt4}.
We consider 102 languages and 606 translation directions (202 English-centric directions, 202 French-centric directions and 202 Chinese-centric directions).
Results show that the multilingual translation capabilities of LLMs are continually involving and GPT-4 reaches new performance height.
Compared with the widely-used supervised MMT system NLLB~\cite{costa2022no}, GPT-4 achieves higher performance on 40.91\% English-centric translation directions.
But compared with the commercial translation system (Google Translate), LLMs still have a long way to go, particularly when it comes to low-resource languages.
French-centric and Chinese-centric translation are also more challenging for GPT-4 than English-centric translation, which further indicates its unbalanced capability across languages.

For the second question, we find some new working patterns.
First, we discover that LLM can acquire translation ability in a resource-efficient way and generate moderate translation even on zero-resource languages.
Second, LLMs are able to perform translation even with unreasonable instructions if in-context learning exemplars are given.
However, if given mismatched translation pairs as in-context exemplars, LLMs fail to translate,
which is similar to observations from concurrent studies~\citep{DBLP:journals/corr/abs-2303-03846}. 
This shows the importance of exemplars in ICL for machine translation. 
Third, we find that cross-lingual translation pairs can be surprisingly good exemplars for low-resource translation, even better than exemplars in the same language.  
The main contribution of this paper can be summarized below:
\begin{itemize}[itemsep=1pt]
    \item We benchmark popular LLMs on MMT in 102 languages and 606 translation directions, covering English-centric, French-centric and Chinese-centric translation.
    \item We systematically compare the results of LLMs and three strong supervised baselines (M2M-100, NLLB, Google Translator) and reveal the gap between two translation paradigms.
    \item We find some new ICL working patterns of LLMs for MMT and discuss corresponding advantages and challenges.
\end{itemize}

\section{Background}
\subsection{Large Language Models}
Language modeling is a long-standing task in natural language processing~\cite{bengio2000neural, mikolov2010recurrent, khandelwal2020generalization}, which is a task to predict the probability of the next token. Transformer~\cite{vaswani2017attention} basically is the backbone of existing LLMs.

LLMs show great potential as a universal multi-task learner. 
Recently, \citet{radford2019language} find that a casual decoder-only language model can be a multi-task learner with merely unsupervised training corpus.
Later, \citet{kaplan2020scaling} reveal the \textit{scaling law} of LLM, indicating that when the scale of neural parameters and training data keeps increasing, LLM can be further strengthened.
\citet{wei2022emergent} show that scaling the language model also brings astonishing \textit{emergent abilities}, e.g., in-context learning, which is only present in large models.
Consequently, more and more efforts have been put into scaling-up language models~\cite{brown2020language, hoffmann2022empirical, scao2022bloom, vilar2022prompting, ren2023pangu}.
Among them, GPT-4~\cite{openai2023gpt4} and ChatGPT~\cite{openai2022chatgpt} are the most representative systems, which show impressive results in various NLP tasks.

\subsection{Emergent Ability: In-context Learning}
In-context learning is one of the well-known emergent abilities~\cite{brown2020language, dong2022survey}, which enables LLM to learn target tasks according to the prompt without updating any parameters.

Specifically, the prompt is made up of in-context exemplars $\{(\mathcal{X}_i,\mathcal{Y}_i)\}_{i=1}^k$ and in-context template $\mathcal{T}$.
Exemplars are often picked from supervised data, where $\mathcal{Y}_i$ is the ground truth corresponding to the input sentence $\mathcal{X}_i$.
Template $\mathcal{T}$ is usually a human-written instruction related to the target task.
Wrapping exemplars with the template and concatenating them together produce the final prompt:
\begin{equation}
\mathcal{P}=\mathcal{T}(\mathcal{X}_1, \mathcal{Y}_1)\oplus\mathcal{T}(\mathcal{X}_2, \mathcal{Y}_2)\oplus\cdots\oplus\mathcal{T}(\mathcal{X}_k, \mathcal{Y}_k)\nonumber
\end{equation}
where $\oplus$ denotes the concatenation symbol, e.g., whitespace, line-break. 
During inference, LLM is able to generate the corresponding output $\mathcal{Y}$ of the test sample $\mathcal{X}$ under the guidance of the prompt:
\begin{equation}
\mathop{\arg\max}_\mathcal{Y} \ p(\mathcal{P} \oplus \mathcal{T}(\mathcal{X}, \mathcal{Y}))
\end{equation}

\begingroup
\renewcommand{\arraystretch}{1.3} % Default value: 1
\begin{table*}[ht]
    \centering
    \footnotesize
    \resizebox{0.95\linewidth}{!}{
    \begin{tabular}{cccccccccccc}
    \toprule
    \multirow{2}{*}{\textbf{Language Family}} & \multirow{2}{*}{\textbf{Direction}} & \multicolumn{10}{c}{\textbf{Translation Performance (BLEU / COMET)}} \\
                                              &                    & XGLM-7.5B     & OPT-175B & Falcon-7B & LLaMA2-7B & LLaMA2-7B-Chat & ChatGPT & GPT-4 & M2M-12B & NLLB-1.3B & Google \\
    \midrule
    \multirow{2}{*}{Indo-Euro-Germanic (8)}    & X$\Rightarrow$Eng & 18.54 / 70.09 & 34.65 / 83.71  & 27.37 / 67.40 & 37.28 / 84.73 & 34.82 / 84.25 & 45.83 / 89.05 & \underline{48.51} / \underline{\textbf{89.48}} & 42.72 / 87.74 & 46.54 / 88.18 & \textbf{51.16} / 89.36 \\
                                               & Eng$\Rightarrow$X & 9.16 / 50.21  & 18.89 / 71.97  &  13.19 / 52.93 & 22.78 / 76.05 & 19.44 / 73.63 & 36.34 / 87.83 & \underline{40.64} / \underline{88.50} & 37.30 / 86.47 & 38.47 / 87.31 & \textbf{45.27} / \textbf{89.05}\\
    \midrule 
    \multirow{2}{*}{Indo-Euro-Romance (8)}     & X$\Rightarrow$Eng & 31.11 / 79.67 & 38.93 / 87.75 & 34.06 / 84.40 & 41.10 / 88.10 & 37.84 / 87.80 & 45.68 / 89.61 & \underline{\textbf{47.29}} / \underline{\textbf{89.74}} & 42.33 / 88.31 & 46.33 / 88.99 & 35.69 / 89.66\\
                                               & Eng$\Rightarrow$X & 21.95 / 69.08 & 24.30 / 79.07  & 20.02 / 70.36 & 27.81 / 82.05 & 25.50 / 79.67 & 41.35 / \underline{\textbf{89.00}} & \underline{\textbf{44.47}} / 88.94 & 42.98 / 87.56 & 43.48 / 88.12 & 37.10 / 88.77\\
    \midrule 
    \multirow{2}{*}{Indo-Euro-Slavic (12)}     & X$\Rightarrow$Eng & 13.20 / 64.24 & 20.83 / 74.80 & 13.15 / 57.34 & 34.00 / 84.90 & 30.94 / 83.90 & 39.27 / 87.74 & \underline{41.19} / \underline{88.15} & 35.87 / 85.97 & 39.23 / 87.08 & \textbf{43.61} / \textbf{88.18}\\
                                               & Eng$\Rightarrow$X & 6.40 / 43.28  & 8.18 / 54.45   & 4.34 / 35.73 & 20.24 / 76.30 & 16.14 / 69.75 & 32.61 / 87.90 & \underline{36.06} / \underline{89.15} & 35.01 / 86.43 & 36.56 / 88.74 & \textbf{42.75} / \textbf{90.05}\\
    \midrule 
    \multirow{2}{*}{Indo-Euro-Indo-Aryan (10)} & X$\Rightarrow$Eng & 8.68 / 63.93  & 1.20 / 49.37  & 1.40 / 45.22 & 6.68 / 62.63 & 4.29 / 60.29 & 25.32 / 84.14 & \underline{37.30} / \underline{87.79} & 17.53 / 69.66 & 40.75 / 88.80 & \textbf{45.66} / \textbf{89.43}\\
                                               & Eng$\Rightarrow$X & 4.76 / 40.99  & 0.14 / 31.85  & 0.13 / 25.84 & 1.61 / 35.92 & 1.24 / 34.74 & 16.50 / 68.43 & \underline{21.35} / \underline{73.75} & 14.44 / 65.32 & 34.04 / 82.55 & \textbf{39.04} / \textbf{82.78}\\
    \midrule 
    \multirow{2}{*}{Indo-Euro-Other (11)}      & X$\Rightarrow$Eng & 7.32 / 55.29  & 7.80 / 59.60  & 7.04 / 51.59 & 14.27 / 69.87 & 11.46 / 67.64 & 29.54 / 84.52 & \underline{37.29} / \underline{86.76} & 22.38 / 77.47 & 36.16 / 86.81 & \textbf{41.68} / \textbf{88.29} \\
                                               & Eng$\Rightarrow$X & 4.51 / 40.60  & 3.10 / 40.04  & 3.38 / 34.64 & 5.00 / 44.09 & 4.83 / 43.73 & 22.81 / 77.33 & \underline{28.45} / \underline{80.94} & 19.71 / 74.90 & 31.65 / 85.82 & \textbf{38.54} / \textbf{87.44}\\
    \midrule 
    \multirow{2}{*}{Austronesian (6)}          & X$\Rightarrow$Eng & 16.19 / 78.80 & 25.60 / 78.03 & 18.62 / 75.36 & 26.70 / 80.21 & 24.39 / 80.39 & 39.95 / 87.29 & \underline{46.81} / \underline{88.65} & 31.84 / 84.76 & 45.41 / 87.85 & \textbf{50.68} / \textbf{88.89}\\
                                               & Eng$\Rightarrow$X & 10.01 / 73.14 & 10.68 / 64.97 &  8.56 / 60.89 & 14.59 / 74.80 & 13.29 / 74.88  & 30.17 / 86.36 & \underline{34.66} / \underline{87.68} & 27.03 / 86.83 & 37.17 / 88.82 & \textbf{40.74} / \textbf{89.34}\\
    \midrule 
    \multirow{2}{*}{Atlantic-Congo (14)}       & X$\Rightarrow$Eng & 6.67 / 62.00  & 9.17 / 57.59  & 6.98 / 0.56 & 8.76 / 57.72 & 9.01 / 57.86 & 19.86 / 79.63 & \underline{28.27} / \underline{83.42} & 10.55 / 76.43 & \textbf{32.20} / 84.00 & 23.55 / \textbf{85.44}\\
                                               & Eng$\Rightarrow$X & 2.52 / 54.93  & 1.60 / 34.15  & 1.89 / 0.34 & 2.45 / 34.17 & 3.09 / 38.13 & 8.91 / 75.26 & \underline{13.70} / \underline{77.79} & 6.53 / 75.79 & \textbf{21.99} / 79.95 & 16.77 / \textbf{80.89} \\
    \midrule 
    \multirow{2}{*}{Afro-Asiatic (6)}          & X$\Rightarrow$Eng & 6.70 / 54.51  & 5.93 / 52.90  & 4.87 / 38.62 & 10.41 / 57.72 & 8.65 / 58.27 & 20.84 / 70.39 & \underline{30.48} / \underline{78.76} & 10.00 / 66.98 & 32.69 / 82.99 & \textbf{36.14} / \textbf{84.47}\\
                                               & Eng$\Rightarrow$X & 2.07 / 41.48  & 1.40 / 41.86  & 1.40 / 27.64 & 3.22 / 43.04 & 3.07 / 43.39 & 13.57 / 67.60 & \underline{19.36} / \underline{75.56} & 7.83 / 68.86 & 26.08 / 82.84 & \textbf{31.00} / \textbf{83.78}\\
   \midrule 
   \multirow{2}{*}{Turkic (5)}                 & X$\Rightarrow$Eng & 7.43 / 61.69  & 7.89 / 62.47  & 4.15 / 33.11 & 9.51 / 65.95 & 8.88 / 66.15 & 24.64 / 84.04 & \underline{31.73} / \underline{86.90} & 10.25 / 58.52 & 32.92 / 87.51 & \textbf{37.78} / \textbf{88.53}\\
                                               & Eng$\Rightarrow$X & 3.48 / 40.32  & 2.58 / 44.80  & 1.75 / 20.00 & 3.28 / 39.65 & 3.09 / 41.97 & 17.13 / 74.77 & \underline{20.96} / \underline{78.50} & 10.87 / 68.21 & 30.17 / 88.47 & \textbf{36.54} / \textbf{89.38}\\
   \midrule 
   \multirow{2}{*}{Dravidian (4)}              & X$\Rightarrow$Eng & 8.04 / 61.95  & 0.89 / 44.01  & 1.18 / 24.29 & 2.65 / 53.17 & 1.52 / 52.95 & 20.26 / 82.00 & \underline{33.10} / \underline{86.91} & 10.26 / 63.77 & 39.07 / 88.42 & \textbf{43.17} / \textbf{89.10}\\
                                               & Eng$\Rightarrow$X & 5.30 / 48.15  & 0.02 / 32.51  & 0.03 / 15.31 & 0.56 / 34.03 & 0.58 / 35.65 & 12.34 / 64.74 & \underline{18.60} / \underline{75.15} & 6.85 / 62.25 & 37.33 / 86.32 & \textbf{44.16} / \textbf{87.75}\\
   \midrule 
   \multirow{2}{*}{Sino-Tibetan (3)}           & X$\Rightarrow$Eng & 9.35 / 58.60  & 9.32 / 65.32  & 16.59 / 72.34 & 18.35 / 74.45 & 16.88 / 74.20 & 21.36 / 78.52 & \underline{27.74} / \underline{84.48} & 11.09 / 71.35 & 30.88 / 86.50 & \textbf{35.68} / \textbf{87.66}\\
                                               & Eng$\Rightarrow$X & 10.14 / 74.16 & 2.57 / 54.73  & 10.74 / 66.74 & 12.24 / 65.99 & 9.06 / 65.07 & 19.92 / 76.04 & \underline{22.81} / \underline{81.11} & 10.42 / 73.82 & 16.85 / 80.74 & \textbf{32.40} / \textbf{88.52}\\
   \midrule 
   \multirow{2}{*}{Other (14)}                 & X$\Rightarrow$Eng & 9.71 / 60.43  & 10.10 / 60.78 & 5.37 / 47.38 & 16.00 / 71.15 & 14.25 / 70.35 & 25.59 / 82.48 & \underline{32.62} / \underline{86.21} & 25.53 / 81.53 & 35.06 / 86.86 & \textbf{36.95} / \textbf{87.93}\\
                                               & Eng$\Rightarrow$X & 8.42 / 51.57  & 3.82 / 46.85  & 1.73 / 29.73 & 8.19 / 53.20 & 7.14 / 52.12 & 20.26 / 74.31 & \underline{24.04} / \underline{79.59} & 23.29 / 77.80 & 28.54 / 85.84 & \textbf{34.34} / \textbf{87.82}\\
    \bottomrule
    \end{tabular}
    }
    \caption{Average translation performance of LLMs on different language families. The number in the bracket indicates the number of evaluated languages in the specific language family. Bold text denotes the highest BLEU or COMET score across models. Underlined text denotes the highest BLEU or COMET score across LLMs.}
    \label{tab:main}
\end{table*}    
\endgroup
\begingroup
\renewcommand{\arraystretch}{1.2} % Default value: 1
\begin{table*}[ht]
    \centering
    \footnotesize
    \resizebox{0.95\linewidth}{!}{
    \begin{tabular}{cccccccccccc}
    \toprule
    \multirow{2}{*}{\textbf{Language Family}} & \multirow{2}{*}{\textbf{Direction}} & \multicolumn{10}{c}{\textbf{Translation Performance (SEScore)}} \\
                                 &                   & XGLM-7.5B & OPT-175B & Falcon-7B & LLaMA-7B & LLaMA-7B-Chat & ChatGPT & GPT4 & M2M-12B & NLLB-1.3B & Google \\
    \midrule
    Indo-Euro-Germanic (8)   & X$\Rightarrow$Eng & -11.78 & -6.00 & -8.34 & -5.41 & -5.90 & -2.52 & \underline{-2.16} & -3.15 & -2.78 & \textbf{-1.85} \\
    Indo-Euro-Romance (8)    & X$\Rightarrow$Eng & -6.54 & -4.01 & -5.57 & -3.72 & -4.14 & -2.30 & \underline{\textbf{-2.08}} & -3.08 & -2.54 & -2.12 \\
    Indo-Euro-Slavic (12)    & X$\Rightarrow$Eng & -14.29 & -10.31 & -13.46 & -5.11 & -5.75 & -3.55 & \underline{-3.17} & -4.21 & -3.70 & \textbf{-2.80} \\
    Indo-Euro-Indo-Aryan (10)& X$\Rightarrow$Eng & -16.45 & -22.15 & -21.65 & -17.15 & -19.46 & -7.64 & \underline{-4.69} & -11.77 & -3.53 & \textbf{-2.80} \\
    Indo-Euro-Other (11)     & X$\Rightarrow$Eng & -18.36 & -17.81 & -18.09 & -13.61 & -15.42 & -6.74 & \underline{-4.62} & -7.57 & \textbf{-3.75} & -4.40 \\
    Austronesian (6)             & X$\Rightarrow$Eng & -14.06 & -10.08 & -12.30 & -9.61 & -10.48 & -4.48 & \underline{-3.03} & -5.37 & -3.47 & \textbf{-2.56} \\
    Atlantic-Congo (14)          & X$\Rightarrow$Eng & -19.42 & -17.61 & -18.44 & -17.59 & -18.48 & -12.38 & \underline{-9.34} & -14.16 & -6.88 & \textbf{-5.75} \\
    Afro-Asiatic (6)             & X$\Rightarrow$Eng & -18.85 & -18.91 & -19.17 & -16.61 & -17.66 & -12.16 & \underline{-8.28} & -14.41 & -4.46 & \textbf{-3.49} \\
    Turkic (5)                   & X$\Rightarrow$Eng & -17.15 & -16.99 & -18.66 & -15.50 & -16.47 & -7.63 & \underline{-5.50} & -15.29 & -4.89 & \textbf{-3.93} \\
    Dravidian (4)                & X$\Rightarrow$Eng & -16.52 & -22.58 & -21.91 & -20.18 & -21.96 & -9.26 & \underline{-5.35} & -13.69 & -3.76 & \textbf{-3.07} \\
    Sino-Tibetan (3)             & X$\Rightarrow$Eng & -19.41 & -15.20 & -12.37 & -11.33 & -12.01 & -10.43 & \underline{-6.79} & -11.93 & -5.50 & \textbf{-4.30} \\
    Other (14)                   & X$\Rightarrow$Eng & -16.74 & -16.56 & -18.70 & -13.05 & -14.17 & -8.51 & \underline{-6.07} & -6.91 & -4.94 & \textbf{-3.80} \\
    \bottomrule
    \end{tabular}
    }
    \caption{Average SEScore of LLMs on different language families. The number in the bracket indicates the number of evaluated languages in the specific language family. Bold text denotes the highest SEScore across models. Underlined text denotes the highest SEScore across LLMs.}
    \label{tab:sescore}
\end{table*}
\endgroup
For label prediction tasks, the prediction $\mathcal{Y}$ can be obtained in one-step generation.
For sequence generation tasks, e.g., machine translation, the prediction $\mathcal{Y}$ can be obtained through sampling strategies like greedy search and beam search.

\section{Experiment Setup}
\noindent\paragraph{Dataset} We benchmark multilingual translation on \textsc{Flores-101}~\cite{goyal2022flores} dataset\footnote{We evaluate LLMs on the first 100 sentences of each direction's test set in benchmarking experiment, considering the prohibitive API cost of evaluating massive languages. In analysis experiment, we use full test set.}, which enables an assessment of model quality on a wide range of languages.

\noindent\paragraph{LLMs} We evaluate translation performance of eight popular LLMs: XGLM-7.5B~\cite{lin2022few}, OPT-175B~\cite{zhang2022opt}, BLOOMZ-7.1B~\cite{scao2022bloom}, Falcon-7B~\cite{falcon40b}, LLaMA2-7B~\cite{touvron2023llama}, LLaMA2-7B-chat~\cite{touvron2023llama}, ChatGPT~\cite{openai2022chatgpt} and GPT-4\footnote{We utilized \textsc{GPT-3.5-Turbo-0301} for ChatGPT (evaluated at April 2023), and \textsc{GPT-4-0613} for GPT-4 (evaluated at August 2023).}~\cite{openai2023gpt4}.

\begin{figure*}
    \centering
    \includegraphics[width=0.9\textwidth]{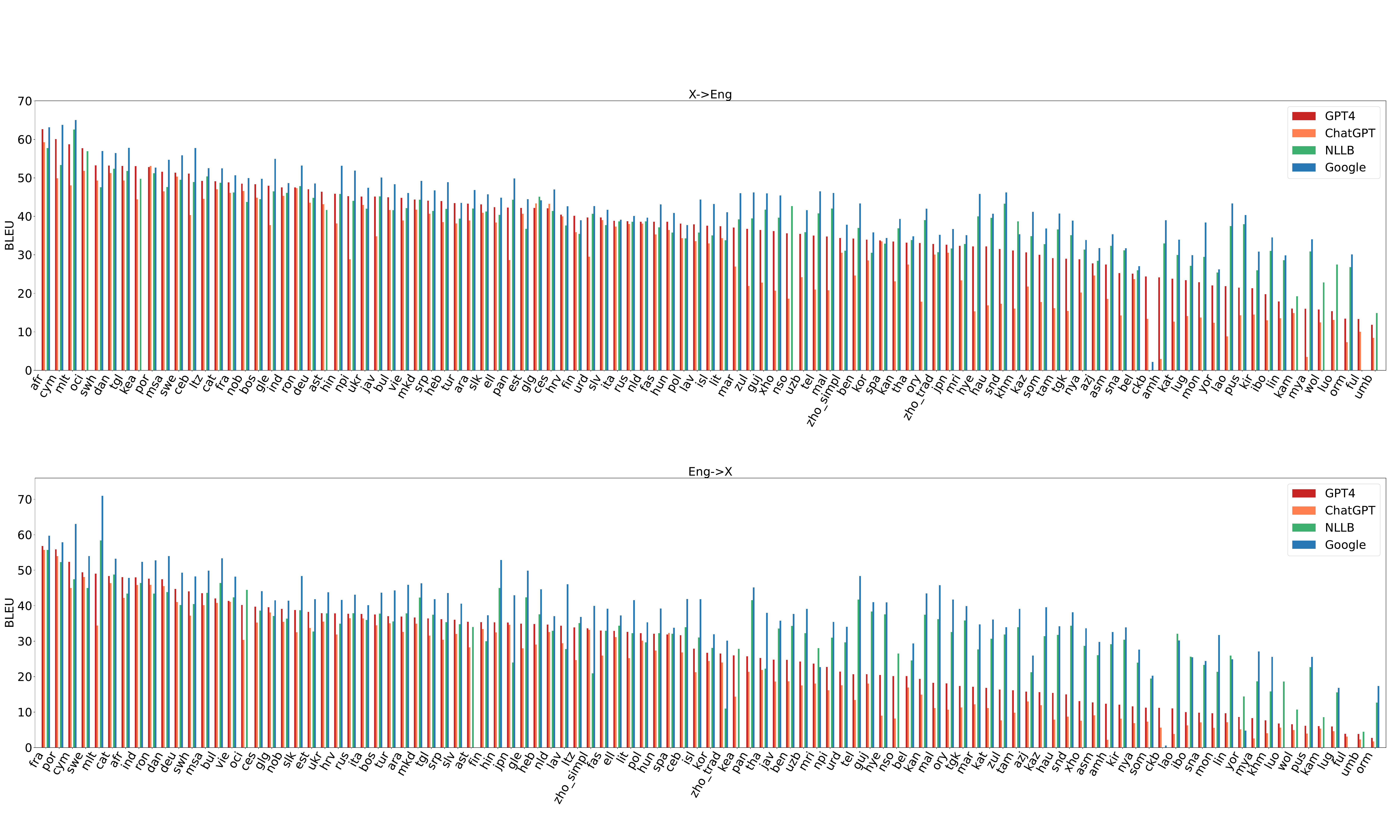}
    \caption{Translation performance (BLEU) of GPT-4, ChatGPT, NLLB and Google Translate on our evaluated languages. ``X->Eng'' and ``Eng->X'' denote translating to English and translating from English respectively. In each subfigure, languages are sorted according to BLEU scores of GPT-4.} 
    \label{fig:chatgpt}
\end{figure*}

\noindent\paragraph{ICL strategy}
For each model, we report its translation performance with eight randomly-picked translation pairs\footnote{We use the same eight randomly-picked translation pairs as exemplars during evaluation.} from the corresponding development set as in-context exemplars and ``<X>=<Y>'' as in-context template. ``<X>'' and ``<Y>'' are the placeholder for the source and target sentence.
We use line-break as the concatenation symbol.
According to our experiment analysis, this ICL strategy serves as a simple but strong recipe.
All implementation is based on \textit{OpenICL}\footnote{\url{https://github.com/Shark-NLP/OpenICL}}~\cite{wu2023openicl}.

\noindent\paragraph{Supervised baselines}
We report the performance of the supervised model M2M-100-12B ~\cite{fan2021beyond} and NLLB-1.3B~\cite{costa2022no} (distillation version), which are widely-used many-to-many MMT models. 
We also report the performance of the powerful commercial translation system, Google Translate\footnote{\texttt{\url{https://translate.google.com/}}}.

\noindent\paragraph{Metric}
Following \citet{goyal2022flores}, we use SentencePiece BLEU\footnote{\url{https://github.com/mjpost/sacrebleu}} (spBLEU) as evaluation metric, which enables an evaluation of all languages.
In addition, we also consider emerging metrics, COMET\footnote{We compute the score with \textit{wmt22-comet-da} model.}~\cite{rei2020comet} and SEScore\footnote{We compute the score with \textit{SEScore-2}~\cite{xu2022sescore2}.}~\cite{xu2022errors}, which have been shown to correlate well with human judgements.

\section{Benchmarking LLMs for Massively  Multilingual Machine Translation}

In this section, we report results on multilingual machine translation and introduce our main findings about LLMs' translation ability.

\noindent\paragraph{The multilingual translation capabilities of LLMs are continually involving}
Table \ref{tab:main} and Table 2\footnote{Currently, SEScore mainly supports evaluating English translation. Thus we evaluate LLM's performance on translating other languages to English.} present evaluation results grouped by language family.
Monolingual pre-trained LLMs present impressive multilingual translation ability, indicating the possibility of aligning multiple languages even with unsupervised data~\cite{garcia2023unreasonable}.
More encouragingly, the multilingual translation capabilities of LLMs are continually improving. 
The most recent LLMs are reaching new performance heights; for example, LLaMA2-7B outperforms previously released open-source LLMs, and GPT-4 surpasses ChatGPT.
Overall, GPT-4 is the best translator among evaluated LLMs and it achieves the highest average BLEU and COMET score on most directions.

\begingroup
\renewcommand{\arraystretch}{1.3} % Default value: 1
\begin{table}[h]
    \centering
    \footnotesize
    \resizebox{0.95\linewidth}{!}{
    \begin{tabular}{ccccccccccccc}
    \toprule
    \textbf{Language Family} & \textbf{X$\Rightarrow$Eng} & \textbf{X$\Rightarrow$Fra} & \textbf{X$\Rightarrow$Zho} & \textbf{Eng$\Rightarrow$X} & \textbf{Fra$\Rightarrow$X} & \textbf{Zho$\Rightarrow$X} &  \\
    \midrule
    Indo-Euro-Germanic (8) & 48.51 & 44.23 & 27.97 & 40.64 & 32.34 & 24.13 \\
    \midrule
    Indo-Euro-Romance (8) & 47.29 & 45.16 & 27.31 & 44.47 & 36.05 & 27.12 \\
    \midrule
    Indo-Euro-Slavic (12) & 41.19 & 40.32 & 25.67 & 36.06 & 30.88 & 23.33 \\
    \midrule
    Indo-Euro-Indo-Aryan (10) & 37.30 & 32.81 & 21.81 & 21.35 & 17.26 & 13.55 \\
    \midrule
    Indo-Euro-Other (11) & 37.29 & 35.36 & 22.70 & 28.45 & 22.57 & 17.50 \\
    \midrule
    Austronesian (6) & 46.81 & 39.98 & 24.40 & 34.66 & 25.64 & 19.52 \\
    \midrule
    Atlantic-Congo (14) & 28.27 & 25.02 & 15.72 & 13.70 & 10.42 & 7.60 \\
    \midrule
    Afro-Asiatic (6) & 30.48 & 27.00 & 17.81 & 19.36 & 14.43 & 10.53 \\
    \midrule
    Turkic (5) & 31.73 & 30.90 & 19.96 & 20.96 & 17.80 & 14.02 \\
    \midrule
    Dravidian (4) & 33.10 & 30.61 & 20.63 & 18.60 & 14.47 & 11.37 \\
    \midrule
    Sino-Tibetan (3) & 27.74 & 27.93 & 20.88 & 22.81 & 19.21 & 16.30 \\
    \midrule
    Other (14) & 32.62 & 31.26 & 21.25 & 24.04 & 20.03 & 16.37 \\
    \bottomrule
    \end{tabular}
    }
    \caption{Translation performance (BLEU) of GPT-4 on English-centric, French-centric and Chinese-centric translation.}
    \label{tab:gpt4}
\end{table}    
\endgroup

\noindent\paragraph{LLM's capability is unbalanced across languages}
In Table~\ref{tab:main}, we observe a similar trend for all evaluated LLMs: they perform better at translating into English than translating into non-English.
LLM's capability on non-English languages is also unbalanced.
For languages that are similar to English, e.g, Indo-European-Germanic languages, LLMs achieve impressive results.
For languages that are dissimilar to English, e.g., Sino-Tibetan languages, LLMs often produce less decent results.

Table~\ref{tab:gpt4} presents another clue, where we evaluate GPT-4 on French-centric and Chinese-centric translation.
Compared to English-centric translation, GPT-4 faces greater challenge when it comes to non-English-centric translation, which again indicates LLM's unbalanced translation ability across languages.

\noindent\paragraph{LLMs still lag behind the strong supervised baseline, especially on low-resource languages}
Figure \ref{fig:chatgpt} shows the translation performance of the supervised systems and GPT-4 on each language.
In 40.91\% translation directions, GPT-4 has achieved higher BLEU scores than NLLB, indicating the promising future of this new translation paradigm.
But on long-tail low-resource languages, GPT-4 still lags behind NLLB, let alone Google Translate.

\begin{figure*}[ht]
    \centering
    \includegraphics[width=0.8\textwidth]{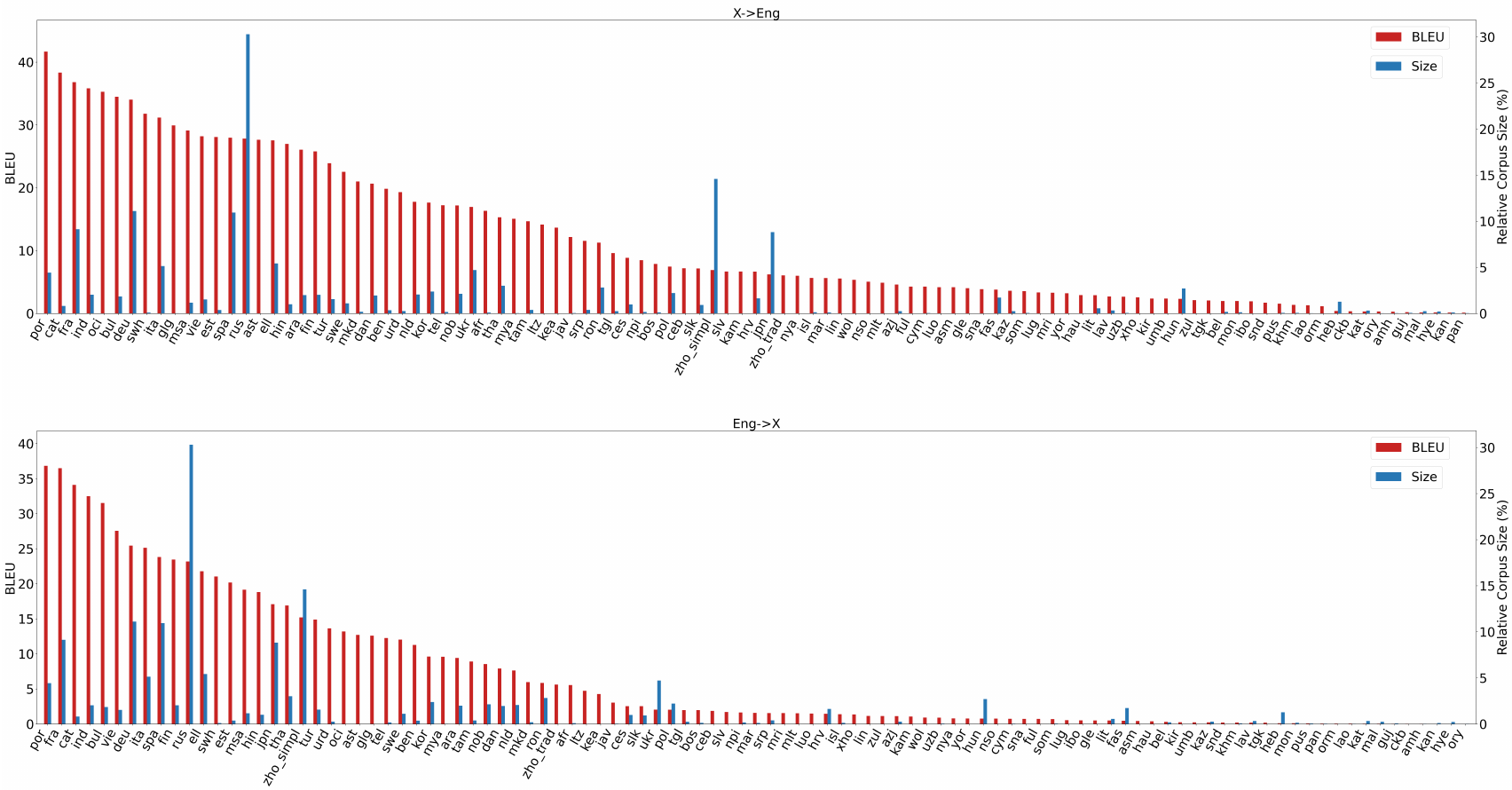}
    \caption{Translation performance (BLEU) of XGLM on evaluated languages and the corpus size of each language relative to English pre-training corpus. In each subfigure, languages are sorted according to BLEU scores of XGLM.}
    \label{fig:corpus}
\end{figure*}

\begin{figure}[ht]
    \centering
    \includegraphics[width=0.40\textwidth]{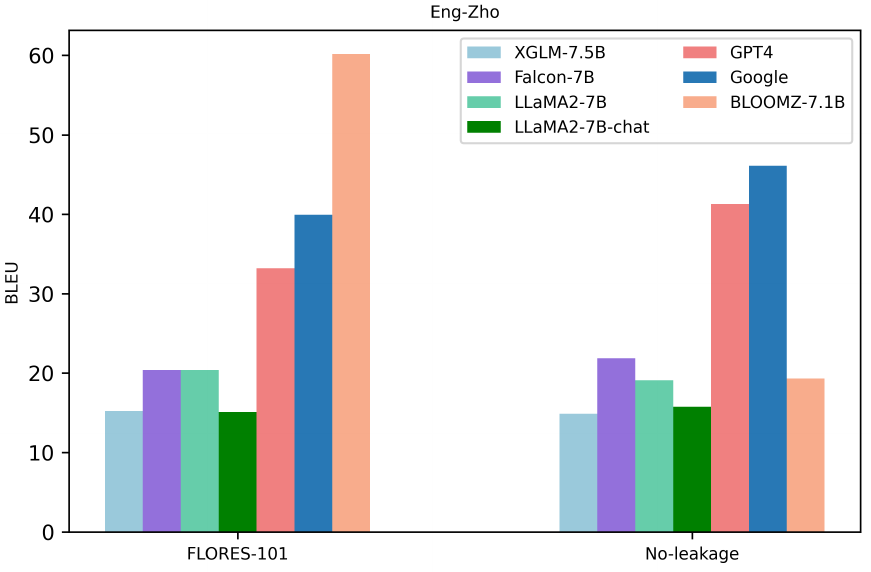}
    \caption{Translation performance of different models on \textsc{Flores-101} test set and our annotated no-leakage evaluation set \textsc{News2023}.}
    \label{fig:no-leakage}
\end{figure}

\noindent\paragraph{Data leakage issue should be considered before evaluating LLMs on public datasets.}
We do not include BLOOMZ's performance on \textsc{Flores-101} in our report because BLOOMZ is instruction-tuned with \textsc{xP3} dataset~\cite{scao2022bloom}, which includes \textsc{Flores-200} dataset. 
Thus BLOOMZ may have been exposed to test cases from \textsc{Flores-101} during training.
If so, the evaluation results can not precisely reflect its translation ability~\cite{elangovan2021memorization}.

To illustrate this concern, we take 1000 English sentences from the most recent news spanning August 2023 to October 2023\footnote{The news were collected from BBC news, Fox news, ABC news and Yahoo news.}, and ask human experts to translate them into Chinese and construct a bilingual no-leakage evaluation set, named \textsc{News2023}.
Figure \ref{fig:no-leakage} shows that BLOOMZ's performance significantly deteriorates on this no leakage set, whereas other models maintain a consistent performance across both datasets.
Through this example, we wish to draw the community's attention to the potential data leakage issue when evaluating large language models.

\section{Analyzing Factors That Influence LLM's Translation Performance }
To better understand how LLM acquires translation ability and which factors have influence on its performance, we conduct in-depth analysis.
For analysis, we choose XGLM-7.5B as an example\footnote{We choose XGLM for three reasons: (1) XGLM has a multilingual focus and covers many languages, which can be seen as a representative of multilingual LLM. (2) XGLM-7.5B is an open-source medium-sized LLM. It is more affordable to run experiments with it than large-sized LLMs or close-source LLMs. (3) The composition of the XGLM's pre-training corpus is clear, allowing us to analyze the relationship between translation ability and corpus size.}.
Note that, when studying a certain factor, we keep the remaining factors unchanged.

\begingroup
\renewcommand{\arraystretch}{1.1} % Default value: 1
\begin{table*}[ht]
    \centering
    \footnotesize
    \scalebox{0.75}{
    \begin{tabular}{cccccccc}
    \toprule    % \multicolumn{2}{c}{\multirow{2}{*}{\textbf{Template}}} & \multicolumn{2}{c}{\textbf{De-En}} &  \multicolumn{2}{c}{\textbf{Ru-En}} & \multicolumn{2}{c}{\textbf{Ru-De}} & \multirow{2}{*}{\textbf{Average}} \\
    \textbf{In-context Template}  & \textbf{Deu-Eng} & \textbf{Eng-Deu} &  \textbf{Rus-Eng} & \textbf{Eng-Rus} & \textbf{Rus-Deu} & \textbf{Deu-Rus} & \textbf{Average} \\
    \midrule
    \multicolumn{8}{l}{\texttt{reasonable instructions: }} \\
    \text{<X>=<Y>}                                        & 37.37 & \textbf{26.49} & 29.66 & 22.25 & 17.66 & \textbf{17.31} & \textbf{25.12} \\
    \text{<X> $\backslash$n Translate from [SRC] to [TGT]: $\backslash$n <Y>} & 37.95 & 26.29 & 29.83 & 20.61 & 17.56 & 15.93 & 24.70 \\
    \text{<X> $\backslash$n Translate to [TGT]: $\backslash$n <Y>} & 37.69 & 25.84 & \textbf{29.96} & 19.61 & 17.44 & 16.48 & 24.50 \\
    \text{<X> $\backslash$n [TGT]: <Y>}                              & 29.94 & 17.99 & 25.22 & 16.29 & 12.28 & 11.71 & 18.91 \\
    \text{<X> is equivalent to <Y>}                              & 23.00 & 4.21 & 17.76 & 9.44 & 8.14 & 9.84 & 12.07 \\
    \text{<X>$\backslash$n can be translated to$\backslash$n <Y>}                              & 37.55 & \textbf{26.49} & 29.82 & 22.14 & 17.48 & 16.40 & 24.98 \\
    \text{[SRC]: <X> $\backslash$n [TGT]: <Y>}                   & 16.95 & 8.90 & 14.48 & 6.88 & 7.86 & 4.01 & 9.85 \\
    \midrule
    \multicolumn{8}{l}{\texttt{unreasonable instructions: }} \\
    \text{<X>$\$$<Y>}                                        & 37.77 & 26.43 & 29.53 & 20.99 & 17.72 & 17.27 & 24.95 \\
    \text{<X> $\backslash$n Translate from [TGT] to [SRC]: $\backslash$n <Y>} & \textbf{38.18} & 26.21 & 29.85 & 20.35 & \textbf{17.75} & 16.63 & 24.83 \\
    \text{<X> $\backslash$n Compile to [TGT]: $\backslash$n <Y>} & 37.39 & 26.35 & 29.68 & 19.91 & 17.52 & 16.15 & 24.50 \\
    \text{<X> $\backslash$n [SRC]: <Y>}                              & 27.86 & 16.69 & 24.41 & 18.16 & 11.98 & 12.60 & 18.62 \\
    \text{<X> is not equivalent to <Y>}                              & 23.50 & 3.92 & 16.90 & 7.80 & 8.06 & 9.23 & 11.57 \\
    \text{<X> $\backslash$n can be summarized as $\backslash$n <Y>}      & 37.46 & 26.24 & 29.42 & \textbf{22.62} & 17.68 & 17.15 & 25.10 \\
    \text{[SRC]: <X> $\backslash$n [SRC]: <Y>}                   & 19.03 & 8.21 & 15.96 & 6.37 & 7.57 & 4.40 & 10.26 \\
    \bottomrule
    \end{tabular}
    }
    \caption{Translation performance (BLEU) of using different templates for in-context learning. The number of in-context exemplars is fixed at eight in this experiment. ``<X>'' and ``<Y>'' denote the placeholder for source and target sentence respectively. ``[SRC]'' and ``[TGT]'' represent the placeholder for source and target language name in English. Bold text denotes the highest score along the column.}
    \label{tab:template}
\end{table*}
\endgroup
\subsection{Findings on Pre-training Corpus Size}
\noindent\paragraph{LLM can acquire translation ability in a resource-efficient way.}
As XGLM authors report data distribution of their pre-training corpus, we can investigate the relationship between translation performance and corpus size (Figure \ref{fig:corpus}).
We find that for low-resource languages, e.g., Catalan (cat) and Swahili (swh), XGLM can generate moderate translation, showing that LLM can build bilingual mapping between non-English and English with a few non-English monolingual resources (less than 1\% of English resources).
Even on unseen languages, e.g., Occitan (oci) and Asturian (ast), XGLM can translate through ICL.
These observations indicate a potential advantage of the novel translation paradigm: LLM can learn to translate in a resource-efficient way.

\subsection{Findings on In-context Template}
\noindent\paragraph{The good performance of LLMs relies on carefully-designed template}
The initial step of applying in-context learning for translation is determining the template.
We find that the translation performance varies greatly with different templates (Table \ref{tab:template}), where the largest gap in the average performance is up to 16 BLEU.
The best template for each direction is also different.
Among these templates, ``<X>=<Y>'' achieves the highest average BLEU score.
``[SRC]: <X> $\backslash$n [TGT]: <Y>'' achieves the lowest score, although it is a commonly-used template for prompting other LLMs, e.g., PaLM~\cite{vilar2022prompting}, GLM~\cite{zhang2023prompting}.
Such phenomena indicate that the template plays a vital role in ICL and it may be challenging to design a universally optimal template for different LLMs and translation directions.

\noindent\paragraph{Even unreasonable template can instruct LLM to generate decent translation}
A common intuition of ICL is that the template instructs LLMs to do the target task~\cite{brown2020language}, e.g., the template ``\text{<X> can be translated to <Y>}'' instructs the LLM to perform translation task.
\begin{figure}[ht]
    \centering
    \includegraphics[width=0.48\textwidth]{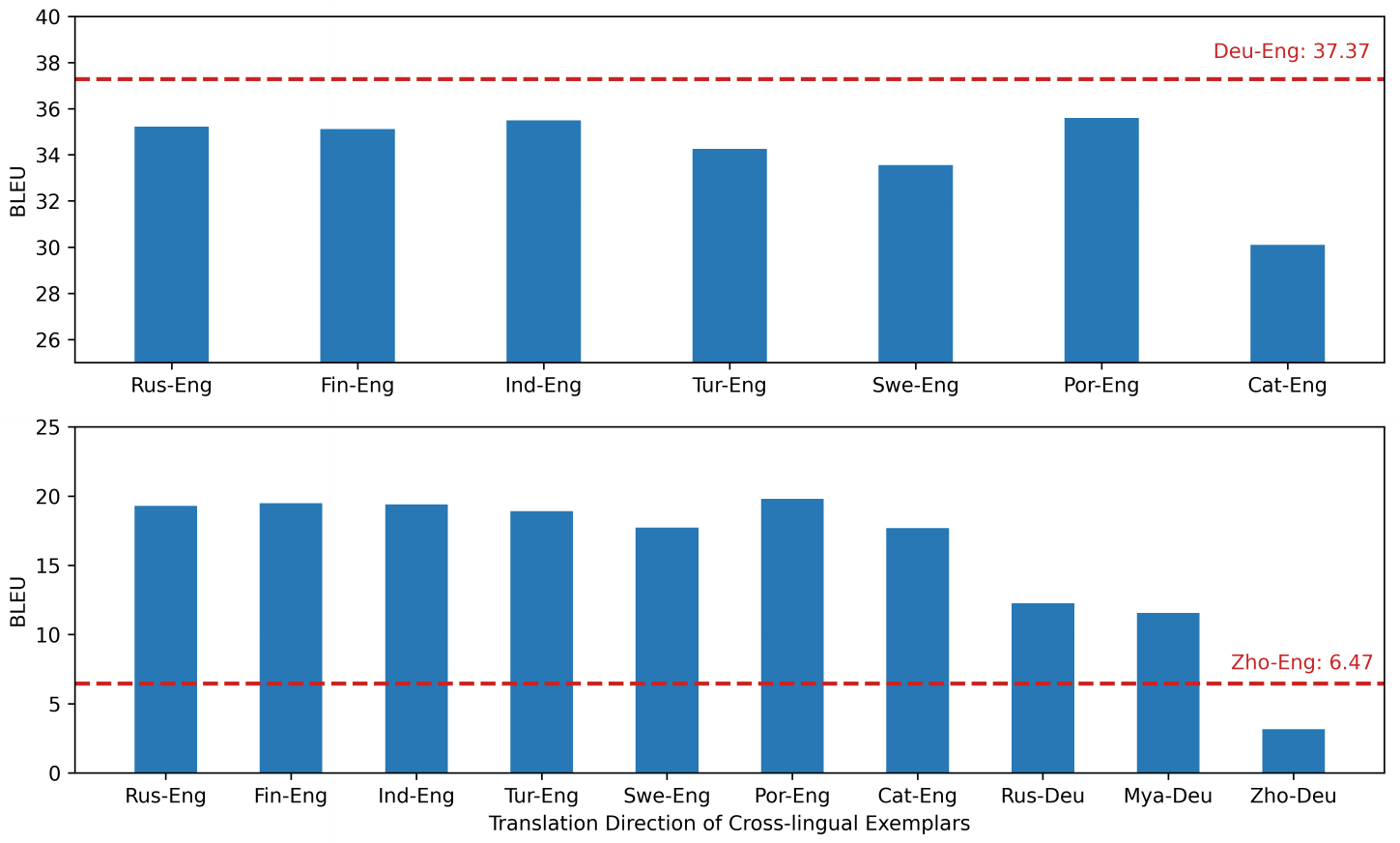}
    \caption{Effects of using cross-lingual exemplars.}
    \label{fig:cross}
\end{figure}
However, we find that wrapping translation exemplars with task-unrelated template can also serve as an effective prompt.
For example, the template like ``\text{<X> can be summarized as <Y>}'' can also instruct LLM to generate translation, rather than guiding it to generate summarization.
Given the fact that these unreasonable template are also effective, the community may not fully understand the role of in-context-template.

\subsection{Findings on In-context Exemplar}
\noindent\paragraph{Cross-lingual exemplars help for certain translation directions}
Translation direction of the exemplar is a unique factor in machine translation.
We find that using cross-lingual exemplars does not always causes worse performance and show two cases in Figure \ref{fig:cross}.
When using cross-lingual exemplars for German-English translation, the translation performance degenerates.
\begingroup
\renewcommand{\arraystretch}{1.25} % Default value: 1
\begin{table*}[ht]
    \centering
    \footnotesize
    \scalebox{0.75}{
    \begin{tabular}{c|ccc|cccc}
    \toprule
    \textbf{In-context Exemplars} & \textbf{Consistency} & \textbf{Granularity} &  \textbf{Diversity} & \textbf{Deu-Eng} & \textbf{Eng-Deu} & \textbf{Zho-Eng} & \textbf{Eng-Zho} \\
    \midrule
    Mismatched Translation & \color{red}\XSolidBrush & \Checkmark & \Checkmark & 0.00 & 0.00 & 0.42 & 1.16 \\
    Word-level Translation & \Checkmark & \color{red}\XSolidBrush & \Checkmark & 25.10 & 5.84 & 2.81 & 2.24 \\
    Doc-level Translation & \Checkmark & \color{red}\XSolidBrush & \Checkmark & 8.01 & 2.05 & 4.48 & 2.20 \\
    Duplicated Translation & \Checkmark & \Checkmark & \color{red}\XSolidBrush & 35.12 & 19.66 & 17.87 & 7.86 \\
    \midrule
    Sent-level Translation  & \Checkmark & \Checkmark & \Checkmark & 37.37 & 26.49 & 19.86 & 11.07 \\
    \bottomrule
    \end{tabular}
    }
    \caption{Translation performance (BLEU) of XGLM when using different contents as in-context exemplars. ``Consistency'' column denotes whether source and target sentence are semantically consistent. ``Granularity'' column denotes whether the exemplar is a sentence-level pair. ``Diversity'' column denotes whether exemplars in the context are different from each other.}
    \label{tab:ablation}
\end{table*}
\endgroup
\begin{figure}[ht]
    \centering
    \includegraphics[width=0.4\textwidth]{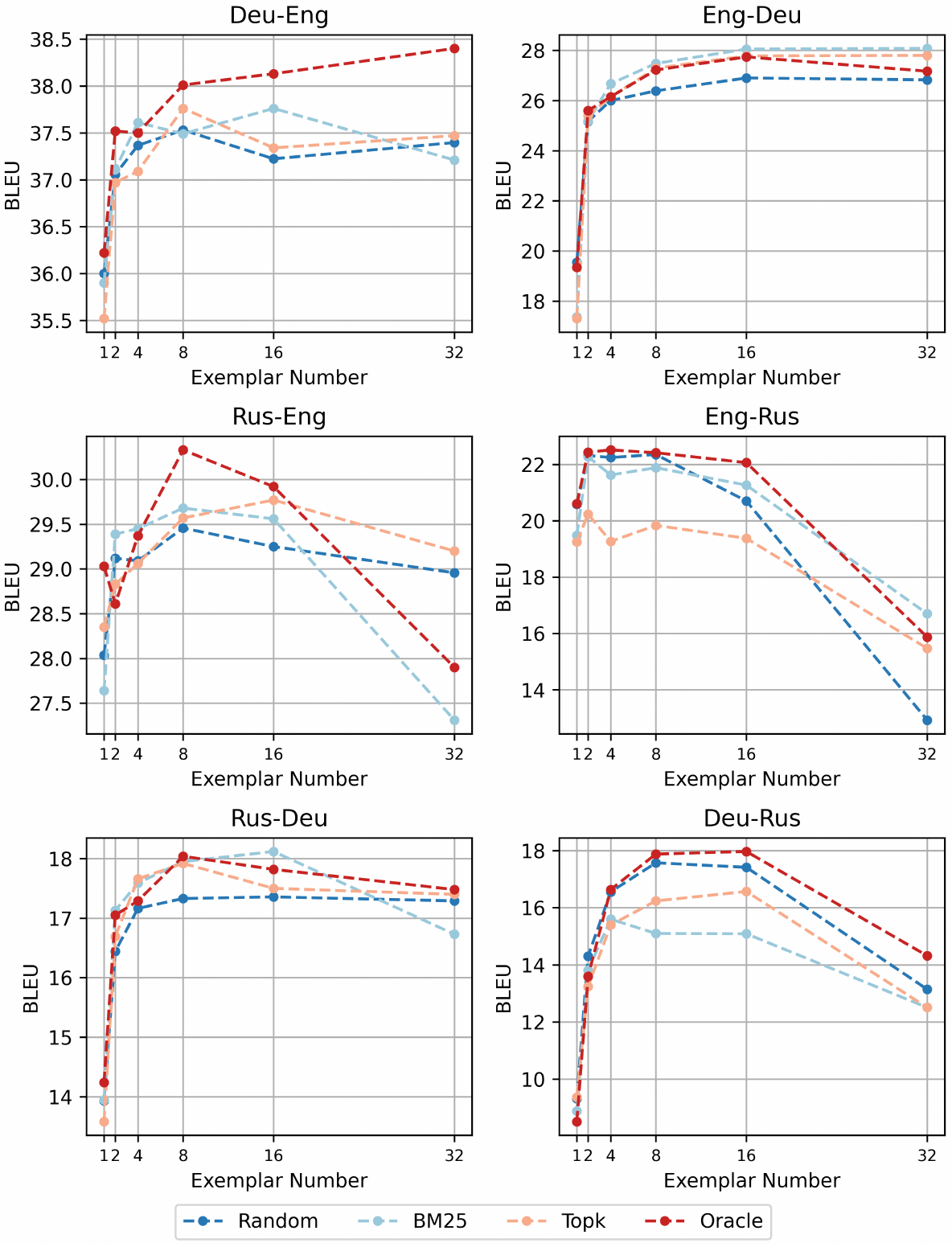}
    \caption{Effects of selecting varying number of in-context exemplars according to different strategies.}
    \label{fig:retriever}
\end{figure}
But when using cross-lingual exemplars for low-resource Chinese-English translation (illustrated in Appendix~\ref{sec:cross}), XGLM's translation performance usually improves significantly, even when both source and target language is changed.
This phenomenon indicates the potential usage of cross-lingual exemplars in a broader range of tasks~\cite{lin2022few}, and we will explore more about this in the future.

\noindent\paragraph{Semantically-related exemplars does not brings more benefits than randomly-picked exemplars}
\begin{table}[ht]
    \centering
    \footnotesize
    \scalebox{0.85}{
    \begin{tabular}{c|cc|cc}
    \toprule
    \textbf{Rev} & \multicolumn{2}{c|}{\textbf{Deu-Eng}} & \multicolumn{2}{c}{\textbf{Eng-Deu}}\\
    \textbf{ratio} &  Head            & Tail &  Head            & Tail \\
    \midrule
    0 / 8            & 37.37 & 37.37 & 26.49 & 26.49 \\ 
    1 / 8            & 37.74 & 36.05 & 26.75 & 23.96 \\ 
    2 / 8            & 37.29 & 36.79 & 26.89 & 24.66 \\ 
    3 / 8            & 36.82 & 35.67 & 26.44 & 24.34 \\ 
    4 / 8            & 36.60 & 35.18 & 26.23 & 22.17 \\ 
    5 / 8            & 35.61 & 31.93 & 25.58 & 17.47 \\ 
    6 / 8            & 30.49 & 20.71 & 22.42 & 8.73 \\ 
    7 / 8            & 14.60 & 5.36  & 12.51 & 3.19 \\ 
    8 / 8            & 3.42 & 3.42   & 3.10 & 3.10 \\ 
    \bottomrule
    \end{tabular}
    }
    \caption{Effects of reversing in-context examples' translation direction. ``Rev ratio'' means the number of exemplars that are reversed. ``Head'' and ``Tail'' represents reversing the exemplars in the head and tail of the prompt respectively.}
    \label{tab:direction}
\end{table}
In this paper, we use development set for exemplar selection, which has been found to be a high-quality candidate pool~\cite{vilar2022prompting}, and we compare four ways of selecting in-context exemplars, namely \textit{Random}\footnote{\textit{Random}: picking exemplars on a random basis.}, \textit{BM25}\footnote{\textit{BM25}: selecting exemplars whose source sentences are similar to the test case's source sentence according to BM25.}, \textit{TopK}\footnote{\textit{TopK}: selecting exemplars whose source sentences are similar to the test case's source sentence according to the similarity of sentence embedding.} and \textit{Oracle}\footnote{\textit{Oracle}: selecting exemplars whose target sentences are similar to the test case's according to sentence embedding, which can be seen as the upper bound of selection strategy.}.

Effects of selecting varying number of in-context exemplars with different approaches are shown in Figure \ref{fig:retriever}.
The general trend in all dataset is similar.
As the number of examples grows from 1 to 8, the BLEU score increases rapidly.
Afterwards, the translation performance plateaus regardless of selection strategy.
When more exemplars are added, e.g., 32 exemplars, the BLEU score usually starts to decline, shows an opposite phenomenon against the observation in natural language understanding tasks~\cite{li2023context}.

Compared to semantically-related exemplars, randomly-picked exemplars gives comparable translation performance.
Even the performance of oracle selection is on par with random selection.
Based on these observations, we suggest that translation exemplars can teach LLM to translate but LLM may struggle to acquire helpful translation knowledge from semantically-related exemplars.

\noindent\paragraph{Exemplars teach LLM the core feature of translation task}
To better understand how ICL exemplars influence LLM to understand the translation task, we observe LLM's translation behaviour under abnormal in-context exemplars (Table \ref{tab:ablation}).

We can see that LLM completely fails when mismatched translation is used as exemplars, indicating that LLM needs to learn from the context to keep source and target sentence semantically consistent.
Word-level\footnote{We select word pairs from open-source \textit{fasttext} dictionary.} and document-level\footnote{We select document translation from Europarl dataset.} translation exemplar degenerates LLM's translation performance, which demonstrates that the translation granularity of exemplar matters as well.
Another interesting phenomenon is that LLM performs worse when duplicated translation is used as the exemplar, indicating that keeping in-context exemplars diverse is also important.
In general, these comparison results show that LLM learns the core feature of translation task through in-context learning.

\noindent\paragraph{The exemplar in the tail of the prompt has more impact on the LLM's behaviour}
During our analysis, we find that reversing the translation direction of exemplars will cause LLM to fail.
Based on this observation, we conduct experiments to investigate the importance of different parts of the prompt (Table \ref{tab:direction}).
We find that reversing exemplars in the tail of the prompt consistently produced worse results compared to reversing exemplars in the head, which suggests that exemplars in the tail of the prompt have larger influence on LLM's behavior.

\section{Related Work}
\noindent\paragraph{In-context learning for machine translation}
Using LLMs for multilingual machine translation is attracting more and more attention.
\citet{lin2022few} evaluate GPT-3 and XGLM-7.5B on 182 directions.
\citet{bawden2023investigating} evaluates BLOOM on 30 directions.
\citet{bang2023multitask}, \citet{jiao2023chatgpt}, \citet{hendy2023good} and \citet{peng2023towards} evaluate ChatGPT on 6 to 18 directions.
In this paper, we thoroughly evaluate multilingual translation performance of popular LLMs on 102 languages and 606 directions and compare them with state-of-the-art translation engines, such as NLLB and Google Translate, which provides a more comprehensive benchmark result and highlights the challenges involved in optimizing this emerging translation paradigm.

To find better ICL recipe for machine translation, many efforts have been put into designing exemplars selection strategy~\cite{agrawal2022context, zhang2023prompting, moslem2023adaptive}. 
Similar to the findings of \citet{zhang2023prompting}, we find that random selection is a simple but effective strategy.
We also find that even oracle selection can not result in consistently better performance.
\citet{wei2022finetuned} shows few-shot exemplars improve translation performance. 
And we further demonstrate the dynamic variations of translation performance with the number of in-context exemplars and the usage of cross-lingual exemplars.
Besides, \citet{vilar2022prompting} find that using a high-quality pool, e.g., development set, for ICL example selection is better and \citet{zhang2023prompting} analyze why the quality of translation exemplars matters.
In this paper, we reveal how in-context exemplars teach LLM to translate by analyzing LLM's behaviour under different kinds of exemplars.

\noindent\paragraph{Multilingual machine translation}
Developing a bilingual translation system for each direction becomes impossible when the number of supporting languages increases.
Therefore, multilingual machine translation is proposed~\cite{johnson2017googles}.
But how to build a high-quality yet efficient MMT system remains an on-going challenge~\cite{costa2022no, yuan2023lego, guerreiro2023hallucinations, robinson2023chatgpt}.
In this paper, we focus on LLM and reveal its potential in MMT.

\section{Conclusion}
In this paper, we evaluate the multilingual translation ability of popular LLMs, including ChatGPT and GPT-4, on 102 languages and 606 directions, which presents the advantages and challenges of LLMs for MMT.
We find that translation capabilities of LLMs are continually involving and GPT-4 reaches new performance height.
However, even for GPT-4, it still face challenge on low-resource languages.
In our analysis, we find that LLMs exhibit new working patterns when used for MMT.
For example, instruction semantics can be ignored during in-context learning and cross-lingual exemplars can provide better task instruction for low-resource translation.
More importantly, we find that LLM can acquire translation ability in a resource-efficient way, which indicates the promising future of LLM in multilingual machine translation.

\section*{Limitations}
In this paper, we mainly evaluate LLM's English-centric, French-centric and Chinese-centric translation ability.
In the future, we would like to investigate more translation directions, e.g., Russian-centric translation, Arabic-centric translation, which could bring more findings concerning with LLM's translation ability.

\section*{Acknowledgement}
We would like to thank Fei Yuan, Zhenyu Wu, Yunzhe Lv for their support to this project.
Shujian Huang is the corresponding author.
This work is partially supported by National Science Foundation of China (No. 62376116, 62176120), the Liaoning Provincial Research Foundation for Basic Research (No. 2022-KF-26-02) and the research project of Nanjing University-China Mobile Joint Institute.

\normalem
% Entries for the entire Anthology, followed by custom entries
\bibliography{anthology,custom}

\begin{thebibliography}{43}
\expandafter\ifx\csname natexlab\endcsname\relax\def\natexlab#1{#1}\fi

\bibitem[{Agrawal et~al.(2022)Agrawal, Zhou, Lewis, Zettlemoyer, and
  Ghazvininejad}]{agrawal2022context}
Sweta Agrawal, Chunting Zhou, Mike Lewis, Luke Zettlemoyer, and Marjan
  Ghazvininejad. 2022.
\newblock In-context examples selection for machine translation.
\newblock \emph{arXiv preprint arXiv:2212.02437}.

\bibitem[{Almazrouei et~al.(2023)Almazrouei, Alobeidli, Alshamsi, Cappelli,
  Cojocaru, Debbah, Goffinet, Heslow, Launay, Malartic et~al.}]{falcon40b}
Ebtesam Almazrouei, Hamza Alobeidli, Abdulaziz Alshamsi, Alessandro Cappelli,
  Ruxandra Cojocaru, Merouane Debbah, Etienne Goffinet, Daniel Heslow, Julien
  Launay, Quentin Malartic, et~al. 2023.
\newblock Falcon-40b: an open large language model with state-of-the-art
  performance, 2023.
\newblock \emph{URL https://huggingface. co/tiiuae/falcon-40b}.

\bibitem[{Bang et~al.(2023)Bang, Cahyawijaya, Lee, Dai, Su, Wilie, Lovenia, Ji,
  Yu, Chung et~al.}]{bang2023multitask}
Yejin Bang, Samuel Cahyawijaya, Nayeon Lee, Wenliang Dai, Dan Su, Bryan Wilie,
  Holy Lovenia, Ziwei Ji, Tiezheng Yu, Willy Chung, et~al. 2023.
\newblock A multitask, multilingual, multimodal evaluation of chatgpt on
  reasoning, hallucination, and interactivity.
\newblock \emph{arXiv preprint arXiv:2302.04023}.

\bibitem[{Bawden and Yvon(2023)}]{bawden2023investigating}
Rachel Bawden and François Yvon. 2023.
\newblock Investigating the translation performance of a large multilingual
  language model: the case of bloom.
\newblock \emph{arXiv preprint arXiv:2303.01911}.

\bibitem[{Bengio et~al.(2000)Bengio, Ducharme, and Vincent}]{bengio2000neural}
Yoshua Bengio, R{\'e}jean Ducharme, and Pascal Vincent. 2000.
\newblock A neural probabilistic language model.
\newblock \emph{Advances in Neural Information Processing Systems (NeurIPS)}.

\bibitem[{Brown et~al.(2020)Brown, Mann, Ryder, Subbiah, Kaplan, Dhariwal,
  Neelakantan, Shyam, Sastry, Askell et~al.}]{brown2020language}
Tom Brown, Benjamin Mann, Nick Ryder, Melanie Subbiah, Jared~D Kaplan, Prafulla
  Dhariwal, Arvind Neelakantan, Pranav Shyam, Girish Sastry, Amanda Askell,
  et~al. 2020.
\newblock Language models are few-shot learners.
\newblock \emph{Advances in Neural Information Processing Systems (NeurIPS)}.

\bibitem[{Dong et~al.(2022)Dong, Li, Dai, Zheng, Wu, Chang, Sun, Xu, Li, and
  Sui}]{dong2022survey}
Qingxiu Dong, Lei Li, Damai Dai, Ce~Zheng, Zhiyong Wu, Baobao Chang, Xu~Sun,
  Jingjing Xu, Lei Li, and Zhifang Sui. 2022.
\newblock A survey for in-context learning.
\newblock \emph{arXiv preprint arXiv:2301.00234}.

\bibitem[{Elangovan et~al.(2021)Elangovan, He, and
  Verspoor}]{elangovan2021memorization}
Aparna Elangovan, Jiayuan He, and Karin Verspoor. 2021.
\newblock Memorization vs. generalization : Quantifying data leakage in {NLP}
  performance evaluation.
\newblock In \emph{Proceedings of the Conference of the European Chapter of the
  Association for Computational Linguistics (EACL)}.

\bibitem[{Fan et~al.(2021)Fan, Bhosale, Schwenk, Ma, El-Kishky, Goyal, Baines,
  Celebi, Wenzek, Chaudhary et~al.}]{fan2021beyond}
Angela Fan, Shruti Bhosale, Holger Schwenk, Zhiyi Ma, Ahmed El-Kishky,
  Siddharth Goyal, Mandeep Baines, Onur Celebi, Guillaume Wenzek, Vishrav
  Chaudhary, et~al. 2021.
\newblock Beyond english-centric multilingual machine translation.
\newblock \emph{The Journal of Machine Learning Research (JMLR)}.

\bibitem[{Garcia et~al.(2023)Garcia, Bansal, Cherry, Foster, Krikun, Feng,
  Johnson, and Firat}]{garcia2023unreasonable}
Xavier Garcia, Yamini Bansal, Colin Cherry, George Foster, Maxim Krikun,
  Fangxiaoyu Feng, Melvin Johnson, and Orhan Firat. 2023.
\newblock The unreasonable effectiveness of few-shot learning for machine
  translation.
\newblock \emph{arXiv preprint arXiv:2302.01398}.

\bibitem[{Goyal et~al.(2022)Goyal, Gao, Chaudhary, Chen, Wenzek, Ju, Krishnan,
  Ranzato, Guzm{\'a}n, and Fan}]{goyal2022flores}
Naman Goyal, Cynthia Gao, Vishrav Chaudhary, Peng-Jen Chen, Guillaume Wenzek,
  Da~Ju, Sanjana Krishnan, Marc{'}Aurelio Ranzato, Francisco Guzm{\'a}n, and
  Angela Fan. 2022.
\newblock The {F}lores-101 evaluation benchmark for low-resource and
  multilingual machine translation.
\newblock \emph{Transactions of the Association for Computational Linguistics
  (TACL)}.

\bibitem[{Guerreiro et~al.(2023)Guerreiro, Alves, Waldendorf, Haddow, Birch,
  Colombo, and Martins}]{guerreiro2023hallucinations}
Nuno~M Guerreiro, Duarte Alves, Jonas Waldendorf, Barry Haddow, Alexandra
  Birch, Pierre Colombo, and Andr{\'e}~FT Martins. 2023.
\newblock Hallucinations in large multilingual translation models.
\newblock \emph{arXiv preprint arXiv:2303.16104}.

\bibitem[{Hendy et~al.(2023)Hendy, Abdelrehim, Sharaf, Raunak, Gabr,
  Matsushita, Kim, Afify, and Awadalla}]{hendy2023good}
Amr Hendy, Mohamed Abdelrehim, Amr Sharaf, Vikas Raunak, Mohamed Gabr, Hitokazu
  Matsushita, Young~Jin Kim, Mohamed Afify, and Hany~Hassan Awadalla. 2023.
\newblock How good are gpt models at machine translation? a comprehensive
  evaluation.
\newblock \emph{arXiv preprint arXiv:2302.09210}.

\bibitem[{Hoffmann et~al.(2022)Hoffmann, Borgeaud, Mensch, Buchatskaya, Cai,
  Rutherford, de~Las~Casas, Hendricks, Welbl, Clark
  et~al.}]{hoffmann2022empirical}
Jordan Hoffmann, Sebastian Borgeaud, Arthur Mensch, Elena Buchatskaya, Trevor
  Cai, Eliza Rutherford, Diego de~Las~Casas, Lisa~Anne Hendricks, Johannes
  Welbl, Aidan Clark, et~al. 2022.
\newblock An empirical analysis of compute-optimal large language model
  training.
\newblock \emph{Advances in Neural Information Processing Systems (NeurIPS)}.

\bibitem[{Jiao et~al.(2023)Jiao, Wang, Huang, Wang, and Tu}]{jiao2023chatgpt}
Wenxiang Jiao, Wenxuan Wang, Jen-tse~Huang Huang, Xing Wang, and Zhaopeng Tu.
  2023.
\newblock Is chatgpt a good translator? yes with gpt-4 as the engine.
\newblock \emph{arXiv preprint arXiv:2301.08745}.

\bibitem[{Johnson et~al.(2017)Johnson, Schuster, Le, Krikun, Wu, Chen, Thorat,
  Vi{\'e}gas, Wattenberg, Corrado, Hughes, and Dean}]{johnson2017googles}
Melvin Johnson, Mike Schuster, Quoc~V. Le, Maxim Krikun, Yonghui Wu, Zhifeng
  Chen, Nikhil Thorat, Fernanda Vi{\'e}gas, Martin Wattenberg, Greg Corrado,
  Macduff Hughes, and Jeffrey Dean. 2017.
\newblock {G}oogle{'}s multilingual neural machine translation system: Enabling
  zero-shot translation.
\newblock \emph{Transactions of the Association for Computational Linguistics
  (TACL)}.

\bibitem[{Kaplan et~al.(2020)Kaplan, McCandlish, Henighan, Brown, Chess, Child,
  Gray, Radford, Wu, and Amodei}]{kaplan2020scaling}
Jared Kaplan, Sam McCandlish, Tom Henighan, Tom~B Brown, Benjamin Chess, Rewon
  Child, Scott Gray, Alec Radford, Jeffrey Wu, and Dario Amodei. 2020.
\newblock Scaling laws for neural language models.
\newblock \emph{arXiv preprint arXiv:2001.08361}.

\bibitem[{Khandelwal et~al.(2020)Khandelwal, Levy, Jurafsky, Zettlemoyer, and
  Lewis}]{khandelwal2020generalization}
Urvashi Khandelwal, Omer Levy, Dan Jurafsky, Luke Zettlemoyer, and Mike Lewis.
  2020.
\newblock Generalization through memorization: Nearest neighbor language
  models.
\newblock In \emph{International Conference on Learning Representations
  (ICLR)}.

\bibitem[{Li et~al.(2023)Li, Gong, Feng, Xu, Zhang, Wu, and
  Kong}]{li2023context}
Mukai Li, Shansan Gong, Jiangtao Feng, Yiheng Xu, Jun Zhang, Zhiyong Wu, and
  Lingpeng Kong. 2023.
\newblock In-context learning with many demonstration examples.
\newblock \emph{arXiv preprint arXiv:2302.04931}.

\bibitem[{Lin et~al.(2022)Lin, Mihaylov, Artetxe, Wang, Chen, Simig, Ott,
  Goyal, Bhosale, Du, Pasunuru, Shleifer, Koura, Chaudhary, O{'}Horo, Wang,
  Zettlemoyer, Kozareva, Diab, Stoyanov, and hLi}]{lin2022few}
Xi~Victoria Lin, Todor Mihaylov, Mikel Artetxe, Tianlu Wang, Shuohui Chen,
  Daniel Simig, Myle Ott, Naman Goyal, Shruti Bhosale, Jingfei Du, Ramakanth
  Pasunuru, Sam Shleifer, Punit~Singh Koura, Vishrav Chaudhary, Brian O{'}Horo,
  Jeff Wang, Luke Zettlemoyer, Zornitsa Kozareva, Mona Diab, Veselin Stoyanov,
  and Xian hLi. 2022.
\newblock Few-shot learning with multilingual generative language models.
\newblock In \emph{Proceedings of the Conference on Empirical Methods in
  Natural Language Processing (EMNLP)}.

\bibitem[{Mikolov et~al.(2010)Mikolov, Karafi{\'a}t, Burget,
  {\v{C}}ernock{\`y}, and Khudanpur}]{mikolov2010recurrent}
Tom{\'a}{\v{s}} Mikolov, Martin Karafi{\'a}t, Luk{\'a}{\v{s}} Burget, Jan
  {\v{C}}ernock{\`y}, and Sanjeev Khudanpur. 2010.
\newblock Recurrent neural network based language model.
\newblock \emph{Interspeech}.

\bibitem[{Moslem et~al.(2023)Moslem, Haque, and Way}]{moslem2023adaptive}
Yasmin Moslem, Rejwanul Haque, and Andy Way. 2023.
\newblock Adaptive machine translation with large language models.
\newblock \emph{arXiv preprint arXiv:2301.13294}.

\bibitem[{OpenAI(2022)}]{openai2022chatgpt}
OpenAI. 2022.
\newblock https://openai.com/blog/chatgpt.

\bibitem[{OpenAI(2023)}]{openai2023gpt4}
OpenAI. 2023.
\newblock \href {http://arxiv.org/abs/2303.08774} {Gpt-4 technical report}.

\bibitem[{Peng et~al.(2023)Peng, Ding, Zhong, Shen, Liu, Zhang, Ouyang, and
  Tao}]{peng2023towards}
Keqin Peng, Liang Ding, Qihuang Zhong, Li~Shen, Xuebo Liu, Min Zhang, Yuanxin
  Ouyang, and Dacheng Tao. 2023.
\newblock Towards making the most of {C}hat{GPT} for machine translation.
\newblock In \emph{Findings of the Association for Computational Linguistics:
  EMNLP 2023}.

\bibitem[{Radford et~al.(2019)Radford, Wu, Child, Luan, Amodei, and
  Sutskever}]{radford2019language}
Alec Radford, Jeffrey Wu, Rewon Child, David Luan, Dario Amodei, and Ilya
  Sutskever. 2019.
\newblock Language models are unsupervised multitask learners.

\bibitem[{Rei et~al.(2020)Rei, Stewart, Farinha, and Lavie}]{rei2020comet}
Ricardo Rei, Craig Stewart, Ana~C Farinha, and Alon Lavie. 2020.
\newblock {COMET}: A neural framework for {MT} evaluation.
\newblock In \emph{Proceedings of Conference on Empirical Methods in Natural
  Language Processing (EMNLP)}.

\bibitem[{Ren et~al.(2023)Ren, Zhou, Meng, Huang, Wang, Wang, Li, Zhang,
  Podolskiy, Arshinov, Bout, Piontkovskaya, Wei, Jiang, Su, Liu, and
  Yao}]{ren2023pangu}
Xiaozhe Ren, Pingyi Zhou, Xinfan Meng, Xinjing Huang, Yadao Wang, Weichao Wang,
  Pengfei Li, Xiaoda Zhang, Alexander Podolskiy, Grigory Arshinov, Andrey Bout,
  Irina Piontkovskaya, Jiansheng Wei, Xin Jiang, Teng Su, Qun Liu, and Jun Yao.
  2023.
\newblock Pangu-$sigma$: Towards trillion parameter language model with sparse
  heterogeneous computing.
\newblock \emph{arXiv preprint arXiv:2303.10845}.

\bibitem[{Robinson et~al.(2023)Robinson, Ogayo, Mortensen, and
  Neubig}]{robinson2023chatgpt}
Nathaniel Robinson, Perez Ogayo, David~R. Mortensen, and Graham Neubig. 2023.
\newblock {C}hat{GPT} {MT}: Competitive for high- (but not low-) resource
  languages.
\newblock In \emph{Proceedings of the Eighth Conference on Machine Translation
  (WMT)}.

\bibitem[{Scao et~al.(2022)Scao, Fan, Akiki, Pavlick, Ili{\'c}, Hesslow,
  Castagn{\'e}, Luccioni, Yvon, Gall{\'e} et~al.}]{scao2022bloom}
Teven~Le Scao, Angela Fan, Christopher Akiki, Ellie Pavlick, Suzana Ili{\'c},
  Daniel Hesslow, Roman Castagn{\'e}, Alexandra~Sasha Luccioni, Fran{\c{c}}ois
  Yvon, Matthias Gall{\'e}, et~al. 2022.
\newblock Bloom: A 176b-parameter open-access multilingual language model.
\newblock \emph{arXiv preprint arXiv:2211.05100}.

\bibitem[{Team(2022)}]{costa2022no}
NLLB Team. 2022.
\newblock No language left behind: Scaling human-centered machine translation.
\newblock \emph{arXiv preprint arXiv:2207.04672}.

\bibitem[{Touvron et~al.(2023)Touvron, Martin, Stone, Albert, Almahairi,
  Babaei, Bashlykov, Batra, Bhargava, Bhosale et~al.}]{touvron2023llama}
Hugo Touvron, Louis Martin, Kevin Stone, Peter Albert, Amjad Almahairi, Yasmine
  Babaei, Nikolay Bashlykov, Soumya Batra, Prajjwal Bhargava, Shruti Bhosale,
  et~al. 2023.
\newblock Llama 2: Open foundation and fine-tuned chat models.
\newblock \emph{arXiv preprint arXiv:2307.09288}.

\bibitem[{Vaswani et~al.(2017)Vaswani, Shazeer, Parmar, Uszkoreit, Jones,
  Gomez, Kaiser, and Polosukhin}]{vaswani2017attention}
Ashish Vaswani, Noam Shazeer, Niki Parmar, Jakob Uszkoreit, Llion Jones,
  Aidan~N Gomez, Lukasz Kaiser, and Illia Polosukhin. 2017.
\newblock Attention is all you need.
\newblock In \emph{Advances in Neural Information Processing Systems
  (NeurIPS)}.

\bibitem[{Vilar et~al.(2022)Vilar, Freitag, Cherry, Luo, Ratnakar, and
  Foster}]{vilar2022prompting}
David Vilar, Markus Freitag, Colin Cherry, Jiaming Luo, Viresh Ratnakar, and
  George Foster. 2022.
\newblock Prompting palm for translation: Assessing strategies and performance.
\newblock \emph{arXiv preprint arXiv:2211.09102}.

\bibitem[{Wei et~al.(2022{\natexlab{a}})Wei, Bosma, Zhao, Guu, Yu, Lester, Du,
  Dai, and Le}]{wei2022finetuned}
Jason Wei, Maarten Bosma, Vincent Zhao, Kelvin Guu, Adams~Wei Yu, Brian Lester,
  Nan Du, Andrew~M Dai, and Quoc~V Le. 2022{\natexlab{a}}.
\newblock Finetuned language models are zero-shot learners.
\newblock In \emph{International Conference on Learning Representations
  (ICLR)}.

\bibitem[{Wei et~al.(2022{\natexlab{b}})Wei, Tay, Bommasani, Raffel, Zoph,
  Borgeaud, Yogatama, Bosma, Zhou, Metzler et~al.}]{wei2022emergent}
Jason Wei, Yi~Tay, Rishi Bommasani, Colin Raffel, Barret Zoph, Sebastian
  Borgeaud, Dani Yogatama, Maarten Bosma, Denny Zhou, Donald Metzler, et~al.
  2022{\natexlab{b}}.
\newblock Emergent abilities of large language models.
\newblock \emph{arXiv preprint arXiv:2206.07682}.

\bibitem[{Wei et~al.(2023)Wei, Wei, Tay, Tran, Webson, Lu, Chen, Liu, Huang,
  Zhou, and Ma}]{DBLP:journals/corr/abs-2303-03846}
Jerry~W. Wei, Jason Wei, Yi~Tay, Dustin Tran, Albert Webson, Yifeng Lu, Xinyun
  Chen, Hanxiao Liu, Da~Huang, Denny Zhou, and Tengyu Ma. 2023.
\newblock Larger language models do in-context learning differently.
\newblock \emph{CoRR}, abs/2303.03846.

\bibitem[{Wu et~al.(2023)Wu, Wang, Ye, Feng, Xu, Qiao, and Wu}]{wu2023openicl}
Zhenyu Wu, YaoXiang Wang, Jiacheng Ye, Jiangtao Feng, Jingjing Xu, Yu~Qiao, and
  Zhiyong Wu. 2023.
\newblock Openicl: An open-source framework for in-context learning.
\newblock \emph{arXiv preprint arXiv:2303.02913}.

\bibitem[{Xu et~al.(2022{\natexlab{a}})Xu, Qian, Wang, Li, and
  Wang}]{xu2022sescore2}
Wenda Xu, Xian Qian, Mingxuan Wang, Lei Li, and William~Yang Wang.
  2022{\natexlab{a}}.
\newblock Sescore2: Retrieval augmented pretraining for text generation
  evaluation.
\newblock \emph{arXiv preprint arXiv:2212.09305}.

\bibitem[{Xu et~al.(2022{\natexlab{b}})Xu, Tuan, Lu, Saxon, Li, and
  Wang}]{xu2022errors}
Wenda Xu, Yi-Lin Tuan, Yujie Lu, Michael Saxon, Lei Li, and William~Yang Wang.
  2022{\natexlab{b}}.
\newblock Not all errors are equal: Learning text generation metrics using
  stratified error synthesis.
\newblock In \emph{Findings of the Association for Computational Linguistics:
  EMNLP 2022}.

\bibitem[{Yuan et~al.(2023)Yuan, Lu, Zhu, Kong, Li, Qiao, and
  Xu}]{yuan2023lego}
Fei Yuan, Yinquan Lu, Wenhao Zhu, Lingpeng Kong, Lei Li, Yu~Qiao, and Jingjing
  Xu. 2023.
\newblock Lego-mt: Towards detachable models in massively multilingual machine
  translation.
\newblock In \emph{Findings of the Association for Computational Linguistics:
  ACL 2023}.

\bibitem[{Zhang et~al.(2023)Zhang, Haddow, and Birch}]{zhang2023prompting}
Biao Zhang, Barry Haddow, and Alexandra Birch. 2023.
\newblock Prompting large language model for machine translation: A case study.
\newblock \emph{arXiv preprint arXiv:2301.07069}.

\bibitem[{Zhang et~al.(2022)Zhang, Roller, Goyal, Artetxe, Chen, Chen, Dewan,
  Diab, Li, Lin et~al.}]{zhang2022opt}
Susan Zhang, Stephen Roller, Naman Goyal, Mikel Artetxe, Moya Chen, Shuohui
  Chen, Christopher Dewan, Mona Diab, Xian Li, Xi~Victoria Lin, et~al. 2022.
\newblock Opt: Open pre-trained transformer language models.
\newblock \emph{arXiv preprint arXiv:2205.01068}.

\end{thebibliography}
\bibliographystyle{acl_natbib}

\appendix

\clearpage
\section{Detailed Results on Each Language}
\label{sec:detail}
\begin{figure*}[ht]
   \centering
    \includegraphics[width=0.90\textwidth]{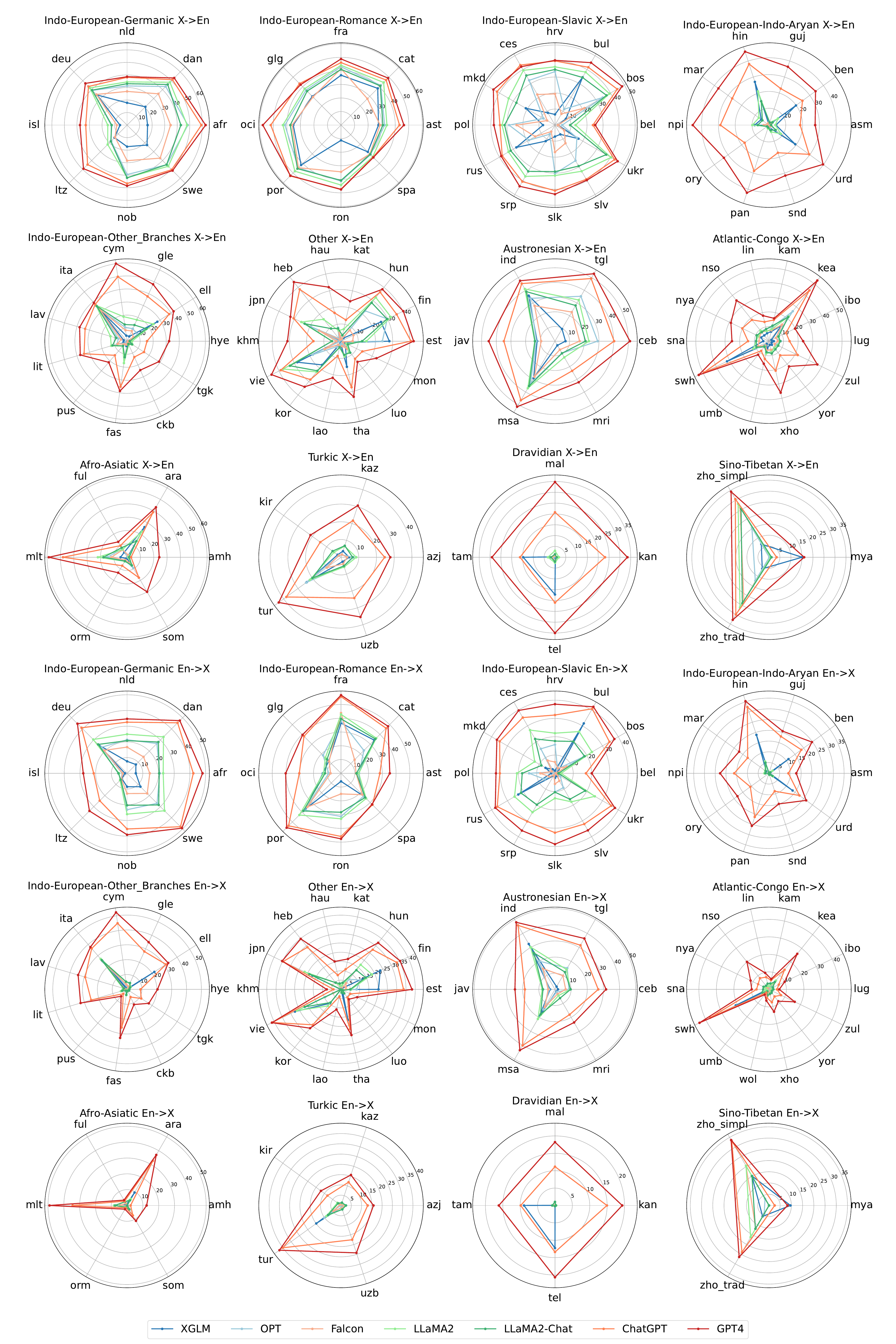}
    \caption{Comparison results (BLEU) between our evalutated LLMs on different language families.}
    \label{fig:6x4}
\end{figure*}

We report detailed results of our evaluated models in Table \ref{tab:all_results_bleu} (BLEU), Table \ref{tab:all_results_comet} (COMET), Table \ref{tab:all_results_sescore} (SEScore) and Figure \ref{fig:6x4}.
One thing that needs to be mentioned is that BLEU supports all translation directions, whereas COMET and SEScore only support a subset of these translation directions.

\section{Lists of Language}
We evaluate 102 languages in this paper.
Table \ref{tab:iso} lists the name, ISO code and language family of these languages.

\section{Cross-lingual Exemplars}
\label{sec:cross}
\begin{figure}[ht]
    \centering
    \includegraphics[width=0.4\textwidth]{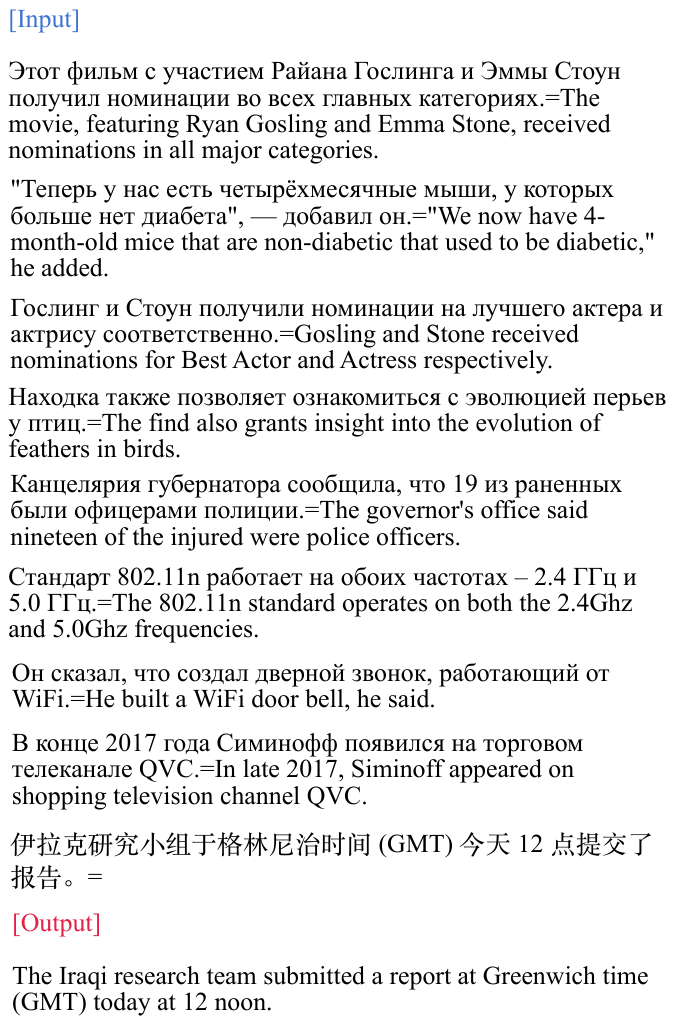}
    \caption{An example of using cross-lingual in-context exemplars}
    \label{fig:cross_case}
\end{figure}

In Figure~\ref{fig:cross_case}, we show an example of using cross-lingual in-context exemplars (Russian-English exemplars for Chinese-English translation).

\section{Used Scientific Artifacts}
Below lists scientific artifacts that are used in our work. For the sake of ethic, our use of these artifacts is consistent with their intended use.
\begin{itemize} [itemsep=1pt]
    \item \textit{OpenICL (Apache-2.0 license)}, a framework that provides an easy interface for in-context learning.
    \item \textit{Transformers (Apache-2.0 license)}, a framework that provides thousands of pretrained models to perform tasks on different modalities such as text, vision, and audio.
\end{itemize}

\begin{table*}[ht]
    \centering
    \footnotesize
    \resizebox{0.95\linewidth}{!}
    {
\begin{tabular}{cc|cccccccccc|cccccccccc}
    \toprule
    \multirow{2}{*}{Language Family} & \multirow{2}{*}{Language} & \multicolumn{10}{c|}{X$\Rightarrow$Eng (BLEU)} & \multicolumn{10}{c}{Eng$\Rightarrow$X (BLEU)} \\
    % \cline{3-7}
                                     &                            & XGLM-7.5B & OPT-175B & Falcon-7B & LLaMA2-7B & LLaMA2-7B-Chat & ChatGPT & GPT4 & M2M-12B & NLLB-1.3B & Google & XGLM-7.5B & OPT-175B & Falcon-7B & LLaMA2-7B &  LLaMA2-7B-Chat & ChatGPT & GPT4 & M2M-12B & NLLB-1.3B & Google\\
    \midrule
    \multirow{8}{*}{Indo-European-Germanic (8)} & afr & 16.34 & 48.49 & 34.73 & 47.89 & 42.89 & 59.28 & 62.65 & 52.86 & 57.76 & 63.15 & 5.56 & 20.75 & 14.45 & 22.98 & 20.42 & 42.18 & 48.02 & 41.41 & 43.39 & 47.83\\
     & dan & 20.65 & 43.54 & 35.31 & 48.33 & 45.83 & 51.23 & 53.18 & 48.32 & 52.35 & 56.44 & 7.91 & 26.81 & 14.80 & 32.79 & 28.19 & 45.49 & 47.46 & 45.12 & 43.81 & 53.99\\
     & nld & 17.78 & 31.25 & 26.87 & 34.46 & 33.03 & 38.10 & 38.60 & 34.52 & 38.68 & 39.66 & 7.64 & 21.38 & 16.69 & 24.89 & 20.80 & 32.57 & 34.66 & 31.79 & 32.93 & 37.05\\
     & deu & 34.03 & 39.15 & 34.60 & 41.94 & 39.44 & 43.56 & 47.04 & 42.79 & 44.79 & 48.52 & 25.44 & 23.38 & 20.65 & 30.46 & 26.01 & 41.02 & 44.69 & 40.18 & 40.20 & 49.32\\
     & isl & 5.65 & 12.68 & 8.18 & 15.41 & 12.28 & 32.98 & 37.58 & 29.47 & 35.07 & 43.19 & 1.40 & 3.10 & 2.77 & 5.13 & 5.53 & 21.26 & 27.89 & 27.80 & 31.04 & 41.80\\
     & ltz & 14.13 & 17.96 & 13.60 & 21.87 & 18.36 & 44.57 & 49.20 & 40.04 & 50.37 & 52.52 & 4.74 & 5.54 & 5.10 & 6.32 & 5.72 & 24.65 & 33.89 & 28.04 & 35.08 & 36.80\\
\bottomrule
    \end{tabular}
    }
    \caption{Detailed results (BLEU) of our evaluated models on 102 languages.}
    \label{tab:all_results_bleu}
\end{table*}

\begin{table*}[ht]
    \centering
    \footnotesize
    \resizebox{0.95\linewidth}{!}
    {
    \begin{tabular}{cc|cccccccccc|cccccccccc}
    \toprule
    \multirow{2}{*}{Language Family} & \multirow{2}{*}{Language} & \multicolumn{10}{c|}{X$\Rightarrow$Eng (COMET)} & \multicolumn{10}{c}{Eng$\Rightarrow$X (COMET)} \\
    % \cline{3-7}
                                     &                            & XGLM-7.5B & OPT-175B & Falcon-7B & LLaMA2-7B & LLaMA2-7B-Chat & ChatGPT & GPT4 & M2M-12B & NLLB-1.3B & Google & XGLM-7.5B & OPT-175B & Falcon-7B & LLaMA2-7B & LLaMA2-7B-Chat & ChatGPT & GPT4 & M2M-12B & NLLB-1.3B & Google\\
    \midrule
    \multirow{7}{*}{Indo-European-Germanic (7)} & afr & 62.96 & 86.21 & 80.54 & 84.93 & 83.45 & 89.87 & 90.33 & 87.24 & 88.36 & 89.73 & 44.67 & 69.54 & 61.85 & 72.62 & 68.30 & 87.00 & 87.48 & 85.64 & 86.56 & 86.94\\
     & dan & 72.74 & 88.46 & 84.15 & 89.10 & 89.04 & 90.54 & 90.67 & 89.74 & 90.03 & 90.76 & 45.95 & 80.09 & 62.31 & 84.14 & 79.85 & 90.79 & 90.45 & 88.89 & 88.92 & 91.02\\
     & nld & 73.32 & 86.41 & 83.69 & 87.42 & 87.27 & 88.62 & 88.64 & 87.42 & 88.04 & 88.39 & 47.82 & 81.52 & 72.40 & 85.26 & 82.11 & 88.87 & 89.03 & 87.43 & 88.39 & 88.97\\
     & deu & 86.13 & 87.75 & 86.71 & 88.08 & 88.30 & 89.39 & 89.61 & 88.48 & 88.98 & 89.50 & 80.23 & 78.06 & 76.54 & 82.88 & 79.99 & 87.95 & 88.51 & 85.61 & 86.26 & 89.13\\
     & isl & 50.24 & 62.09 & 54.35 & 66.68 & 64.91 & 85.35 & 87.13 & 83.43 & 85.25 & 87.50 & 29.53 & 32.46 & 32.01 & 37.53 & 42.71 & 79.22 & 83.28 & 81.51 & 82.92 & 87.16\\
     & nob & 69.85 & 86.82 & 81.47 & 87.55 & 87.55 & 89.25 & 89.47 & 88.08 & 86.94 & 89.07 & 48.40 & 81.45 & 64.72 & 83.68 & 80.21 & 89.86 & 89.86 & 87.58 & 88.37 & 89.67\\
\bottomrule
    \end{tabular}
    }
    \caption{Detailed results (COMET) of our evaluated models.}
    \label{tab:all_results_comet}
\end{table*}

\begin{table*}[ht]
    \centering
    \footnotesize
    \resizebox{0.7\linewidth}{!}
    {
    \begin{tabular}{cc|cccccccccc}
    \toprule
    \multirow{2}{*}{Language Family} & \multirow{2}{*}{Language} & \multicolumn{10}{c}{X$\Rightarrow$Eng (SEScore)} \\
                                     &                            & XGLM-7.5B & OPT-175B & Falcon-7B & LLaMA2-7B & LLaMA2-7B-Chat & ChatGPT & GPT4 & M2M-12B & NLLB-1.3B & Google \\
    \midrule
    \multirow{8}{*}{Indo-European-Germanic (8)} & afr & -13.42 & -3.14 & -6.72 & -3.59 & -4.30 & -0.48 & -0.19 & -1.77 & -1.67 & -0.24\\
     & dan & -10.98 & -3.18 & -5.76 & -2.66 & -2.68 & -1.35 & -1.15 & -1.98 & -1.65 & -0.93\\
     & nld & -10.66 & -4.76 & -6.31 & -4.25 & -4.53 & -3.66 & -3.59 & -4.17 & -3.59 & -3.41\\
     & deu & -4.44 & -3.41 & -4.56 & -3.17 & -3.23 & -2.21 & -2.04 & -2.74 & -2.37 & -1.93\\
     & isl & -20.12 & -15.12 & -18.05 & -13.19 & -14.49 & -4.99 & -4.09 & -5.59 & -5.04 & -3.16\\
     & ltz & -12.83 & -11.46 & -13.68 & -10.39 & -11.89 & -3.14 & -2.12 & -4.06 & -2.01 & -1.52\\
\bottomrule
    \end{tabular}
    }
    \caption{Detailed results (SEScore) of our evaluated models.}
    \label{tab:all_results_sescore}
\end{table*}

% Involved languages in this paper are listed in Table .
\begin{table*}[ht]
    \centering
    \footnotesize
    \resizebox{0.95\linewidth}{!}
    {
    \begin{tabular}{cccc|cccc}
    \toprule
    Language & ISO 639-1 & ISO 639-2/T & Language family & Language & ISO 639-1 & ISO 639-2/T & Language family \\
    \midrule
    Afrikaans & af & afr & Indo-European-Germanic & Latvian & lv & lav & Indo-European-Other \\
    Amharic & am & amh & Afro-Asiatic & Lingala & ln & lin & Atlantic-Congo \\
    Arabic & ar & ara & Afro-Asiatic & Lithuanian & lt & lit & Indo-European-Other \\
    Armenian & hy & hye & Indo-European-Other & Luo & luo & luo & Other \\
    Assamese & as & asm & Indo-European-Indo-Aryan & Luxembourgish & lb & ltz & Indo-European-Germanic \\
    Asturian & ast & ast & Indo-European-Romance & Macedonian & mk & mkd & Indo-European-Slavic \\
    Azerbaijani & az & azj & Turkic & Malay & ms & msa & Austronesian\\
\bottomrule
    \end{tabular}
    }
    \caption{For each language, we list its language name, ISO code and language family.}
    \label{tab:iso}
\end{table*}

\end{document}